\documentclass[11pt,a4paper]{article}
\usepackage[hyperref]{emnlp-ijcnlp-2019}
\usepackage{times}
\usepackage{latexsym}
\usepackage{color}
\usepackage{bm}
\usepackage{latexsym}
\usepackage{amssymb,amsmath}
\usepackage{mathrsfs}
\usepackage{graphicx}
\usepackage{caption}
\usepackage{url}
\usepackage{caption}
\usepackage{tabularx}

\aclfinalcopy % Uncomment this line for the final submission
\usepackage{booktabs}
\usepackage{mathrsfs}
\usepackage{makecell}
\usepackage{multirow}
\usepackage[colorinlistoftodos,prependcaption,textsize=tiny]{todonotes}
\usepackage{enumitem}
\usepackage{hyperref}
\usepackage{algorithm}
\usepackage[noend]{algpseudocode}

\usepackage{xspace}
\newcommand{\metric}[1]{\textsc{#1}\xspace}
\newcommand{\R}{\mathbb{R}}

\newcommand{\1}{\boldsymbol{1}}

\newcommand{\bh}{\boldsymbol{h}}
\newcommand{\bv}{\boldsymbol{v}}

\newcommand{\bz}{\boldsymbol{z}}
\newcommand{\bx}{\boldsymbol{x}}
\newcommand{\by}{\boldsymbol{y}}
\newcommand{\f}{\boldsymbol{f}}

\newcommand{\T}{\mathsf{T}}

\def\mC{{\bm{C}}}

\def\mF{{\bm{F}}}

\def\ervx{{\textnormal{x}}}
\def\ervy{{\textnormal{y}}}

\newtheorem{prop}{Proposition}

\newcommand{\softmax}{\mathit{softmax}}
\DeclareMathOperator*{\argmax}{arg\,max}

\newcommand{\bert}{\textsc{BERT}\xspace}
\newcommand{\bertbase}{\textsc{BERT}_{{\rm base}}\xspace}
\newcommand{\elmobase}{\textsc{ELMo}_{{\rm original}}\xspace}
\newcommand{\methodp}{P_\bert}
\newcommand{\methodr}{R_\bert}
\newcommand{\methodf}{F_\bert}
\newcommand{\idf}{{\rm idf}}
\newcommand{\WMD}{\text{WMD}}
\newcommand{\SMD}{\text{SMD}}

\newcommand\numberthis{\addtocounter{equation}{1}\tag{\theequation}}

\algrenewcommand\algorithmicforall{\textbf{foreach}}
\algrenewcommand\algorithmicindent{.8em}

\usepackage[font={small}]{caption}
\title{MoverScore: Text Generation Evaluating with Contextualized Embeddings and Earth Mover Distance}
\author{Wei Zhao$^{\dagger}$, Maxime Peyrard$^{\dagger}$, Fei Liu$^{\ddagger}$, Yang Gao$^{\dagger}$, Christian M. Meyer$^{\dagger}$, Steffen Eger$^{\dagger}$\\
{$^\dagger$ Computer Science Department, Technische Universit\"at Darmstadt, Germany}\\
{$^{\ddagger}$ Computer Science Department, University of Central Florida, US}\\
{\tt zhao@aiphes.tu-darmstadt.de, maxime.peyrard@epfl.ch} \\
{\tt feiliu@cs.ucf.edu, yang.gao@rhul.ac.uk} \\
{\tt meyer@ukp.informatik.tu-darmstadt.de}\\ 
{\tt eger@aiphes.tu-darmstadt.de}
}

\date{}

\begin{document}

\maketitle
\begin{abstract}

A robust evaluation metric has a profound impact on the development of text generation systems. 
A desirable metric compares system output against references based on their semantics rather than surface forms.
In this paper we investigate strategies to encode system and reference texts to devise a metric that shows a high correlation with human judgment of text quality.
We validate our new metric, namely MoverScore, on a number of text generation tasks including summarization, machine translation, image captioning, and data-to-text generation, where the outputs are produced by a variety of neural and non-neural systems.
Our findings suggest that metrics combining contextualized representations with a distance measure perform the best.
Such metrics also demonstrate strong generalization capability across tasks. 
For ease-of-use we make our metrics available as web service.\footnote{Our code is publicly available at \href{http://tiny.cc/vsqtbz}{http://tiny.cc/vsqtbz}}

\end{abstract}

\section{Introduction}\label{sec:introduction}

The choice of evaluation metric has a significant impact on the assessed quality of natural language outputs generated by a system.
A desirable metric assigns a single, real-valued score to the system output by comparing it with one or more reference texts for content matching.
Many natural language generation (NLG) tasks can benefit from robust and unbiased evaluation, including text-to-text (\emph{machine translation} and \emph{summarization}), data-to-text (\emph{response generation}), and image-to-text (\emph{captioning})~\cite{Gatt:2018}.
Without proper evaluation, it can be difficult to judge on system competitiveness, hindering the development of advanced algorithms for text generation.
%\todo{SE: feel free to make comments like this} 

It is an increasingly pressing priority to develop better evaluation metrics given the recent advances in neural text generation.
% \todo{ChM: debatable -- inventing a new metric just for the sake of it, is not helpful, since it becomes increasingly hard to compare system results if everybody uses her/his own metric. I suggest pointing out the issues with currently used evaluation metrics first (i.e., what is the problem that needs to be solved with a new metric?) FL: I assume researchers from these fields (summarization, MT, NLG) have a consensus that the standard metrics (ROUGE and BLEU) don't work well.. and I agree things won't work if everyone uses their own metrics. Our goal here is to find a metric that can achieve high correlations with human judgments in all conditions, such as clean and corrupted outputs, various generation tasks, neural and non-neural systems etc. This paper is an effort towards this direction but it hasn't addressed all problems.}{}\todo{ChM: I agree that people working the field know that, still I suggest to point out the issues a bit more prominently. How about replacing `new evaluation metrics' with `improved/better evaluation metrics' in the previous sentence and adding `which is hardly/not covered by existing metrics' to the following sentence? FL: Sounds good. I changed the text.}{}
Neural models provide the flexibility to copy content from source text as well as generating unseen words~\cite{See:2017}.
This aspect is hardly covered by existing metrics.
With greater flexibility comes increased demand for unbiased evaluation. 
Diversity-promoting objectives make it possible to generate diverse natural language descriptions~\cite{Li:2016:NAACL,Wiseman:2018}.
But standard evaluation metrics including BLEU~\cite{Papineni:2002} and ROUGE~\cite{Lin:2004} compute the scores based primarily on n-gram co-occurrence statistics,
%\todo{Wei: updated}
which are originally proposed for diagnostic evaluation of systems
% are friendly to diagnostic evaluation 
but not capable of evaluating text quality~\cite{reiter-2018-structured}, as
they are not designed to measure if, and to what extent, the system and reference texts with distinct surface forms have conveyed the same meaning.
% \yg{Recent study
% \cite{chaganty:2018,DBLP:conf/emnlp/no_reference19} has shown that these metrics have low correlation with
% human judgements at sentence level; hence, 
% training NLG systems by hill-climbing on these metrics 
% often leads to poor performance in terms of human scores.}
%\todo{YG: I add these works as I find they further motivate this work. Wei: updated}
Recent effort on the applicability of these metrics reveals that while compelling text generation system ascend on standard metrics, the text quality of system output is still hard to be improved \cite{DBLP:conf/emnlp/no_reference19}.

% Standard evaluation metrics including BLEU~\cite{Papineni:2002} and ROUGE~\cite{Lin:2004} are insufficient to characterize the \emph{semantic relatedness} between system and reference texts, as such metrics are primarily based on computing n-gram co-occurrence statistics between system and reference.\todo{SE: are we all clear on the semantic relatedness issues? Word embeddings capture many things, from grammar, syntax, morphology, to semantics. Is it only important to capture relatedness? Also some people distinguish between semantic similarity and relatedness}

Our goal in this paper is %thus 
to devise an automated evaluation metric assigning a single holistic score to any system-generated text by comparing it against human references for content matching.
% \todo{ChM: what means human reference for semantic relatedness? is this a new type of dataset? or do we conduct a user study in which users also indicate semantic relatedness of system and reference texts? FL: the comment is a bit unclear, I changed the text to content matching but not sure if it helps. ChM: I added the comment, since the text could be misinterpreted as: (1) our new metric is able to match system- and human-generated texts at the level of content/semantics and we use standard datasets with reference texts to check for this. (2) We evaluate the metric's ability with a different (new??) type of dataset that specially captures content matching/semantic relatedness, i.e. that provides the association strength of individual words or content units as annotated by humans, rather than datasets that contain reference texts. After re-reading, I admit that this is likely a minor detail as it become clear later anyway, so feel free to delete the comment.}{} 
We posit that it is crucial to provide a holistic measure attaining high correlation with human judgments so that various neural and non-neural text generation systems can be compared directly. 
Intuitively, the metric assigns a perfect score to the system text if it conveys the same meaning as the reference text. 
Any deviation from the reference content can then lead to a reduced score, e.g., the system text contains more (or less) content than the reference, or the system produces ill-formed text that fails to deliver the intended meaning.

%The challenge lies in building semantic representations for system and reference texts such that their distance accurately reflects any deviation of the system text from its reference.
We investigate the effectiveness of a spectrum of distributional semantic representations to encode system and reference texts, allowing them to be compared for semantic similarity across %a variety of 
multiple 
natural language generation tasks.
% \todo{SE: Kusner is WMD and the other two are contextualizers. Do they go together or should they be separated as done below? FL: removed WMD.}
Our new metric quantifies the semantic distance between system and reference texts by harnessing the power of contextualized representations~\cite{Peters:2018,Devlin:2018} and a powerful distance metric~\cite{Rubner2000} for better content matching. %\todo{SE: ELMO+BERT cited twice in a short distance. Suggestion: delete the first one, since we also use static embeddings. FL: done. nice catch.}
Our contributions %of this paper 
can be summarized as follows: 
\begin{itemize}[topsep=5pt,itemsep=0pt,leftmargin=*]
    
\item We formulate the problem of evaluating generation systems as measuring the semantic distance between system and reference texts, assuming powerful continuous representations can encode any type of semantic and syntactic deviations.

% \item we conduct extensive experiments to understand to what extent existing contextualized representations can accomplish this goal and what it takes to improve,
\item 
% Through extensive experiments, 
We investigate the effectiveness of existing contextualized representations and Earth Mover's Distance~\cite{Rubner2000} for %measuring %semantic distance, 
comparing system predictions and reference texts, 
%\todo{SE: what it takes to improve?? Wei: updated} 
% \todo{ChM: here's something missing. mention that we propose a new metric. FL: I might change the wording after the metric and results are finalized. FL: I updated the wording}{} 
leading to our new 
% an 
automated evaluation metric that achieves high correlation with human judgments of text quality.
%\todo[inline]{SE: how about this one: We propose a new metric }
%\todo[inline]{SE: I don't know how to fix that, but it does not speak about WMD, only about contextualized word embeddings, like the last bullet point. Here's a plausible fix, but more thoughts may be given to this: We perform extensive experiments, showing that contextualized representations and WMD }
%\todo{SE: I think our contributions, ranging from WMD to contextualized embeddings}

\item Our metric outperforms or performs comparably to
% compares favorably to\todo{ChM: what means favorably? is it just on par or does it yield clear improvements?. FL: on par with, but I might change this wording after seeing the final results.}{} 
strong baselines on four text generation tasks including summarization, machine translation, image captioning, and data-to-text generation, suggesting this is a promising direction moving forward. %\todo{SE: for future research? Also: `including' suggests that we do more, but I think the enumeration is already exhaustive}

\end{itemize}

\section{Related Work}\label{sec:related}

% In NLG tasks, like summarization, machine translation, data-to-text generation, image captioning, and many others, systems produce textual outputs which are evaluated in comparison to human-written references. Most previous works in the evaluation of NLG tasks have been carried out independently by each area. This resulted in task-specific metrics despite the shared goal of measuring semantic similarity. Before introducing our general metrics working for a variety of NLG tasks, we briefly present the standard evaluation metrics used in the NLG task that we consider.

It is of fundamental importance to design evaluation metrics that can be applied to natural language generation tasks of similar nature, including summarization, machine translation, data-to-text generation, image captioning, and many others. 
All these tasks involve generating texts of sentence or paragraph length.
The system texts are then compared with one or more reference texts of similar length for semantic matching, whose scores indicate how well the systems perform on each task.
In the past decades, however, evaluation of these natural language generation tasks has largely been carried out independently within each area.

\vspace{0.05in}
\noindent\textbf{Summarization} \quad
A dominant metric for summarization evaluation is ROUGE~\cite{Lin:2004}, which measures the degree of lexical overlap between a system summary and a set of reference summaries.
Its variants consider overlap of unigrams (\textsf{\small -1}), bigrams (\textsf{\small -2}), unigrams and skip bigrams with a maximum gap of 4 words (\textsf{\small -SU4}), longest common subsequences (\textsf{\small -L}) and its weighted version (\textsf{\small -W-1.2}), among others.
% Even though these metrics have become standard, they also lack the ability to capture semantic variations. This becomes particularly problematic with the progress of abstractive summarization. 
% Thus, \newcite{ng-abrecht:2015:EMNLP} extended ROUGE with word embeddings. Instead of hard lexical matching of n-grams, ROUGE-WE uses soft matching based on the cosine similarity of word embedding. However, ROUGE-WE does not correlate with human judgment significantly better than ROUGE. We recommand \citep{Lloret:2013} for a recent and detailed account of the progress in evaluation of automatic summaries.
% {ROUGE scores achieved good correlations with human judgments 
% at system level, i.e.\ by aggregating ROUGE scores of system  
% summaries %’  ROUGE  scores  
% across  multiple input documents,  
% we can reliably rank summarization systems by their quality.
% However, ROUGE performs poorly at summary level: 
% given multiple summaries for the same input document, 
% ROUGE can hardly distinguish good summaries from
% mediocre and bad ones \cite{chaganty:2018,DBLP:conf/emnlp/no_reference19}.}
%in the past.
%\todo{SE: did they? I thought that everyone criticizes them because they do not. YG: I added a few sentences to address SE's comment}
Metrics such as Pyramid~\cite{nenkova-passonneau:2004:HLTNAACL} and BE~\cite{hovy:2006automated,tratz2008summarization} further compute matches of content units, e.g., (head-word, modifier) tuples, that often need to be manually extracted from reference summaries. 
% \todo{YG: The above discussion about Pyramid should not be put here, because the 'However' sentence below does not apply to Pyramid. I think we can either delete the Pyramid part, or put to the end of the paragraph.}
% {\citet{LouisN13} propose several metrics that do not rely on the 
% reference summaries but instead only take
% the summary and the input document(s) as input.
% However, these metrics correlate
% worse with human scores than ROUGE.}
%\todo{YG: the non-reference based metrics are also worth mentioning}
These metrics achieve good correlations with human judgments in the past.
% More recently, \citet{peyrard-2019-simple} 
% measure the importance of summary a composite score combining redundancy, relevance and informativeness to improve summary evaluation.
However, they
% all metrics above %\todo{YG: these $\to$ all metrics above. Wei: updated} 
are not general enough to account for the %semantic 
relatedness between abstractive summaries and their references, as a system abstract can convey the same meaning using different surface forms.
Furthermore, large-scale summarization datasets such as CNN/Daily Mail~\cite{Hermann:2015} and Newsroom~\cite{Grusky:2018} use a \emph{single reference} summary, making it harder to obtain unbiased results when only lexical overlap is considered during summary evaluation.

\vspace{0.05in}
\noindent\textbf{Machine Translation}\quad
A number of metrics are commonly used in MT evaluation.
Most of these metrics compare system and reference translations based on surface forms such as word/character n-gram overlaps and edit distance, but not the meanings they convey.
BLEU~\cite{Papineni:2002} is a precision metric measuring how well a system translation overlaps with human reference translations using n-gram co-occurrence statistics.
Other metrics include SentBLEU, NIST, chrF, TER, WER, PER, CDER, and METEOR~\cite{Lavie:2007} that are used and described in the WMT metrics shared task~\cite{Bojar:2017,Ma:2018}.
RUSE~\cite{Shimanaka:2018} is a recent effort to improve MT evaluation by training sentence embeddings on large-scale data obtained in other tasks. 
% HUME~\cite{birch-etal-2016-hume} is an alternative human judgment in MT compared to direct assessment~\cite{graham-etal-2013-continuous}, which decomposes system and reference texts over UCCA \cite{abend-rappoport-2013-universal} semantic units and provide a composite score by multifaceted judgments on these units.
Additionally, preprocessing reference texts is crucial in MT evaluation, e.g., normalization, tokenization, compound splitting, etc.
If not handled properly, different preprocessing strategies can lead to inconsistent results using word-based metrics~\cite{Post:2018}.

\vspace{0.05in}
\noindent\textbf{Data-to-text Generation}\quad
BLEU can be poorly suited to evaluating data-to-text systems such as dialogue response generation and image captioning.
These systems are designed to generate texts with lexical and syntactic variation, communicating the same information in many different ways.
BLEU and similar metrics tend to reward systems that use the same wording as reference texts, causing repetitive word usage that is deemed undesirable to humans~\cite{Liu:2016:EMNLP}.
In a similar vein, evaluating the quality of image captions can be challenging. 
CIDEr~\cite{VedantamZP15} uses tf-idf weighted n-grams for similarity estimation; 
and SPICE~\cite{AndersonFJG16} incorporates synonym matching over scene graphs.
Novikova et al.~\shortcite{novikova-etal-2017-need} examine a large number of word- and grammar-based metrics and demonstrate that they only weakly reflect human judgments of system outputs generated by data-driven, end-to-end natural language generation systems. 

\vspace{0.05in}
\noindent\textbf{Metrics based on Continuous Representations}\quad
Moving beyond traditional metrics, we envision a new generation of automated evaluation metrics comparing system and reference texts based on semantics rather than surface forms to achieve better correlation with human judgments. 
A number of previous studies exploit static word embeddings ~\cite{ng-abrecht:2015:EMNLP,Lo17} and trained classifers ~\cite{Peyrard:2017, Shimanaka:2018} to improve semantic similarity estimation, replacing lexical overlaps.
% ~\cite{ng-abrecht:2015:EMNLP,Lo17,Peyrard:2017}
%\todo{SE: RUSE also? What is the problem with them? Why do they not perform well? Because t hey use static embeddings or trained classifiers? Wei: updated}

In contemporaneous work, Zhang et al.~\shortcite{zhang:2019} describe a method comparing system and reference texts for semantic similarity leveraging the BERT representations~\cite{Devlin:2018}, which can be viewed as a special case of our metrics and will be discussed in more depth later. 
%\todo{YG: later in section 3. Better to specify later in where}  %\todo{SE: maybe mention here that we will compare to them later on and that they are CONTEMPORANEOUS! Wei: fixed} 
% In similarly contemporaneous work,
More recently,
\citet{clark-etal-2019-sentence} present a semantic metric relying on sentence mover's similarity and the ELMo representations ~\cite{Peters:2018} 
%, which hold the promise of better summary evaluation in multi-sentence contexts,  
and apply them to summarization and essay scoring.
\citet{mathur-etal-2019-putting} introduce unsupervised and supervised metrics based on the BERT representations to improve MT evaluation, %\todo{SE: as a reader, I'm interested to know where the differences are? Do we evaluate on more tasks? Is our WMD better? Do we explore more options? Better aggregating? Or is it more or less the same and people came up ten times with the same idea---this would not be our fault and totally ok. Wei: updated}
% \todo{SE: I still don't know what are the main differences, but maybe none of us does. We probably all don't understand the subtle details of these different approches.}
while \citet{peyrard-2019-simple} provides a composite score combining redundancy, relevance and informativeness to improve summary evaluation. 

% measure the importance of summary a composite score combining redundancy, relevance and informativeness to improve summary evaluation.

In this paper, we seek to accurately measure the (dis)similarity between system and reference texts drawing inspiration from contextualized representations and Word Mover's Distance (WMD; Kusner et al., 2015)\nocite{Kusner:2015}.
WMD finds the ``traveling distance'' of moving from the word frequency distribution of the system text to that of the reference, which is essential to capture the (dis)similarity between two texts. Our metrics differ from the contemporaneous work in several facets: (i) we explore the granularity of embeddings, leading to two variants of our metric, word mover and sentence mover; (ii) we investigate the effectiveness of diverse pretrained embeddings and finetuning tasks; (iii) we study the approach to consolidate layer-wise information within contextualized embeddings;
(iii) our metrics demonstrate strong generalization capability across four tasks, oftentimes outperforming the supervised ones. 
%In the following 
We now describe our method in detail.

\vspace{-0.059in}
\section{Our MoverScore Meric}
We have motivated the need for better metrics capable of evaluating disparate NLG tasks.
We now describe our metric, namely MoverScore, built upon a combination of (i) contextualized representations of system and reference texts and (ii) a distance between these representations measuring the semantic distance between system outputs and references.
It is particularly important for a metric to not only capture the amount of \emph{shared content} between two texts, i.e., \textsf{\small intersect(A,B)}, as is the case with many semantic textual similarity measures~\cite{Peters:2018,Devlin:2018};
but also to accurately reflect to what extent the system text has \emph{deviated} from the reference, i.e., \textsf{\small union(A,B)} - \textsf{\small intersect(A,B)}, which is the intuition behind using a distance metric.
% \todo{SE: I don't understand this. Does it have a real meaning? There's the obvious equation: Similarity $=$ 1-Distance. Does WMD account for distance somewhere?}
% our metrics.\todo{SE: where does that appear? In the WMD part? FL: that's a good catch. I changed the text to be "intuition behind using a distance metric."}
\subsection{Measuring Semantic Distance}
% \subsection{Contextualized Word Mover's Distance}
% A well-known issue is that word-overlap metrics are biased
% and correlate poorly with human judgements of text quality. There are many obvious cases where these metrics fail [], they are often incapable of considering the semantic similarity between system output and human-written references. Despite this, recent work still use them to evaluate texts [] since there are few alternatives available that correlate highly with human judgements.
Let $\bx = (\ervx_1, \dots, \ervx_m)$ be a sentence viewed as a sequence of words.
%\todo{SE: but you use set notation, which means: unordered and no repeated elements.} 
We denote by $\bx^n$ the sequence %list 
of $n$-grams of $\bx$ (i.e., $\bx^1=\bx$ is the sequence %list 
of words and $\bx^2$ is the sequence of bigrams).
% In this section, $S_x^n$ denotes the lists of $n$-grams of a sentence $x$, and $|S_x^n|$ the number of $n$-grams in $x$.
Furthermore, let $\f_{\bx^n}\in\R_+^{|\bx^n|}$ be a %set 
%sequence 
vector 
of weights, %, one 
%for
%corresponding to 
%each 
one weight for each 
%the 
$n$-gram of $\bx^n$. %and w.l.o.g. 
We can assume $\f_{\bx^n}^{\T}\1=1$, making $\f_{\bx^n}$ a distribution over $n$-grams. Intuitively, the effect of some $n$-grams like those including function words can be downplayed by giving them lower weights, e.g., using Inverse Document Frequency (IDF).
% Then, $x = (S_x^n, \f_x)$ represents the system's output and $y = (S_y^n, \f_y)$ is the human reference.

Word Mover's Distance (WMD)~\cite{Kusner:2015}, a special case of Earth Mover's Distance~\cite{Rubner2000}, measures semantic distance between texts by aligning semantically similar words and finding 
% the corresponding travel costs.
the amount of flow traveling between these words.
% \todo{SE: these travel costs, are they the `distance' part? Wei: updated}
%of them. \todo{SE: aligning similar words? BERTscore also does that... Wei: fixed} 
It was shown %to be 
useful for text classification and textual similarity tasks~\cite{Kusner:2015}.
Here, we formulate a generalization operating on $n$-grams.
Let $\bx$ and $\by$ be two sentences viewed as sequences of $n$-grams: $\bx^n$ and $\by^n$. If we have a distance metric $d$ between $n$-grams, then we can define the \emph{transportation cost} matrix $\mC$ such that $\mC_{ij} = d(\ervx^n_i, \ervy^n_j)$ is the distance between the $i$-th $n$-gram of $\bx$ and the $j$-th $n$-gram of $\by$. %\todo{SE: $d$ or $\fd$. In math, I learned that $x$ and $\mathbf{x}$ are two different symbols and should not be confused, for any $x$. This appears below also .... YG: good point, should be $d$ I guess. I read the rest of the subsection and find Wei consistently uses boldface for vectors and non-bold for element. The $\fd$ here is the only exception. Wei: thanks. updated}
The $\WMD$ between the two sequences 
%\todo{SE: you use the words "sets", "sequences", "lists", "vectors" apparently completely interchangeable, but they have precise mathematical definitions.}{} 
of $n$-grams $\bx^n$ and $\by^n$ with associated $n$-gram weights $\f_{\bx^n}$ and $\f_{\by^n}$ is then given by:
\begin{align*}
& \WMD(\bx^n,\by^n) := \min_{\mF \in\R^{|\bx^n|\times |\by^n|}} \langle \mC, \mF \rangle, \\
&\text{s.t. } \mF\1=\f_{\bx^n},\;\;\mF^{\intercal}\1=\f_{\by^n}.
\end{align*}
where $\mF$ is the \emph{transportation flow} matrix with $\mF_{ij}$ denoting the amount of flow traveling from the $i$-th $n$-gram $\ervx^n_i$ in $\bx^n$ to the $j$-th $n$-gram $\ervy^n_j$ in $\by^n$. 
Here, $\langle \mC, \mF \rangle$ denotes the sum of all matrix entries of the matrix $\mC\odot \mF$, where $\odot$ denotes element-wise multiplication. 
%, i.e., Hadamard-product. 
% , and $C$ is the associated transportation cost: $C_{ij}:=\text{dist}(\bv_{x_i},\bv_{y_j})$ given by the distance between two $n$-grams.
% For example, $C_{ij}$ could be given by the cosine distance between the embedding representations of $\bv_{x_i}$ and $\bv_{y_j}$.
Then $\WMD(\bx^n,\by^n)$ is the minimal transportation cost between $\bx^n$ and $\by^n$ where $n$-grams are weighted by $\f_{\bx^n}$ and $\f_{\by^n}$.

% \paragraph{Analysis on WMD}
% The BERScore can be reformulated using WMD notations, defined as:
% \begin{align*}
% \rm{BERTScore-R} (\bx^1, \by^1)  &:= \sum_{i}c_i \cdot f_i \\
% &= \sum_{i}IDF(x_i^1)\max_{y_i^1\in \by^1}E(x_i^1)^\intercal E(y_i^1)
% \end{align*}

In practice, we compute the Euclidean distance between the embedding representations of $n$-grams: 
% to compute the distance: 
$d(\ervx^n_i, \ervy^n_j) = ||E(\ervx^n_i) - E(\ervy^n_j)||_2$ where 
%\todo{YG: if you change $\fd$ to $d$ above, remember to change here also. Wei: thanks. updated}
% $\cos$ is the cosine similarity and 
$E$ is the embedding function which maps an $n$-gram to its vector representation. Usually, \emph{static} word embeddings like word2vec are used to compute $E$ but these cannot capture word order or compositionality. Alternatively, we investigate contextualized embeddings like ELMo and BERT because they encode information about the whole sentence into each word vector.

We compute the $n$-gram embeddings as the weighted sum over its word embeddings.
Formally, if $\bx^n_i = (\ervx_i, \dots, \ervx_{i+n-1})$ is the $i$-th $n$-gram from sentence $\bx$, its embedding is given by:
\begin{equation}
\label{eq:n-gram-embedding}
E(\bx^n_i) = \sum_{k=i}^{i+n-1} \idf(\ervx_k)\cdot E(\ervx_k)
\end{equation}
where $\idf(\ervx_k)$ is the IDF of word $\ervx_k$ computed from all sentences in the corpus and $E(\ervx_k)$ is its word vector.
Furthermore, the weight associated to the $n$-gram 
$\bx^n_i$ is given by:
\begin{equation}
\f_{\bx^n_i} = \frac{1}{Z} \sum_{k=i}^{i+n-1} \idf(\ervx_k)
\end{equation}
where $Z$ is a normalizing constant s.t. $\f_{\bx^n}^{\T}\1=1$,

%Note that, 
In the limiting case where $n$ is larger than the sentence's size, $\bx^n$ contains only one $n$-gram: the whole sentence. Then $\WMD(\bx^n,\by^n)$ reduces to computing the distance between the two sentence embeddings, namely Sentence Mover's Distance (SMD), denoted as:
\begin{align*}
\SMD(\bx^n,\by^n) := ||E(\bx_1^{\l_x})-E(\by_1^{\l_y})||
\end{align*}
where $\l_{x}$ and $\l_{y}$ are the size of sentences. 
%Word and Sentence Mover's Distance are crucial to modeling the semantic distance between system and reference texts, especially for 
% Word Mover Distance %, which 
% solves a constrained optimization problem %and the score indicates 
% to find the minimum amount of effort required to transform between two texts via a soft distance measure based on embeddings.\footnote{The formulation also provides an important possibility to bias the metric towards precision or recall by using an asymmetric cost matrix. We will discuss this later.} \todo{SE: now WMD sounds like soft edit distance. Maybe move this text up}
% Any deviation of the system text from its reference, i.e., containing more or less content, can lead to an increased distance score. 
%\todo{SE: this paragraph needs small fixes. I don't think it's good to talk about EMD here - it was never really introduced. I think putting this paragraph ealier is also better. Wei: I made small changes}

\paragraph{Hard and Soft Alignments}\quad 
% \todo{SE: how about doing this later? Wei: i feel it's ok}
In contemporaneous work, BERTScore \citep{zhang:2019} also %suggests 
models 
the semantic distance between system and reference texts  
%are crucial 
for 
evaluating text generation systems. 
As shown in Figure \ref{fig:bertscore-wmd}, 
% In this section, we compare them in a 
% a method comparing system and reference texts for semantic similarity, which is similar to
% WordMover and BERTScore are used to measure the semantically textual similarity for a sentence pair
% Intuitively, 
BERTScore (precision/recall) can be intuitively viewed as hard alignments (one-to-one) for words in a sentence pair, where each word in one sequence %would 
travels to the most semantically similar word in the other sequence. In contrast, 
% Mover Distance
MoverScore
goes beyond BERTScore as it relies on soft alignments (many-to-one) and allows to map semantically related words in one sequence to the respective word in the other sequence by solving a constrained optimization problem: finding the minimum %amount of 
effort %required 
to transform between two texts.
% \todo{SE: I think BERTScore allows many units to associate with the same unit. The difference between soft and hard is not one-to-one vs.\ many-to-one, but arbitrary probability distribution vs.\ uniform distribution. Wei: C in MoverScore is symmetric (eclidean distance matrix) while C in BERTScore is asymmetric. See appendix. SE: it remains true that BERTScore allows several units to align with the same unit. BERTScore is not one-to-one.}
%Any deviation of the system text from its reference, i.e., containing more or less content, can lead to an increased distance score. 

%We also find that 
The formulation of Word Mover's Distance provides an important possibility to bias the metric towards precision or recall by using an asymmetric transportation cost matrix, which bridges a gap between MoverScore and BERTScore:
\begin{figure}
\centering
\includegraphics[width=\linewidth]{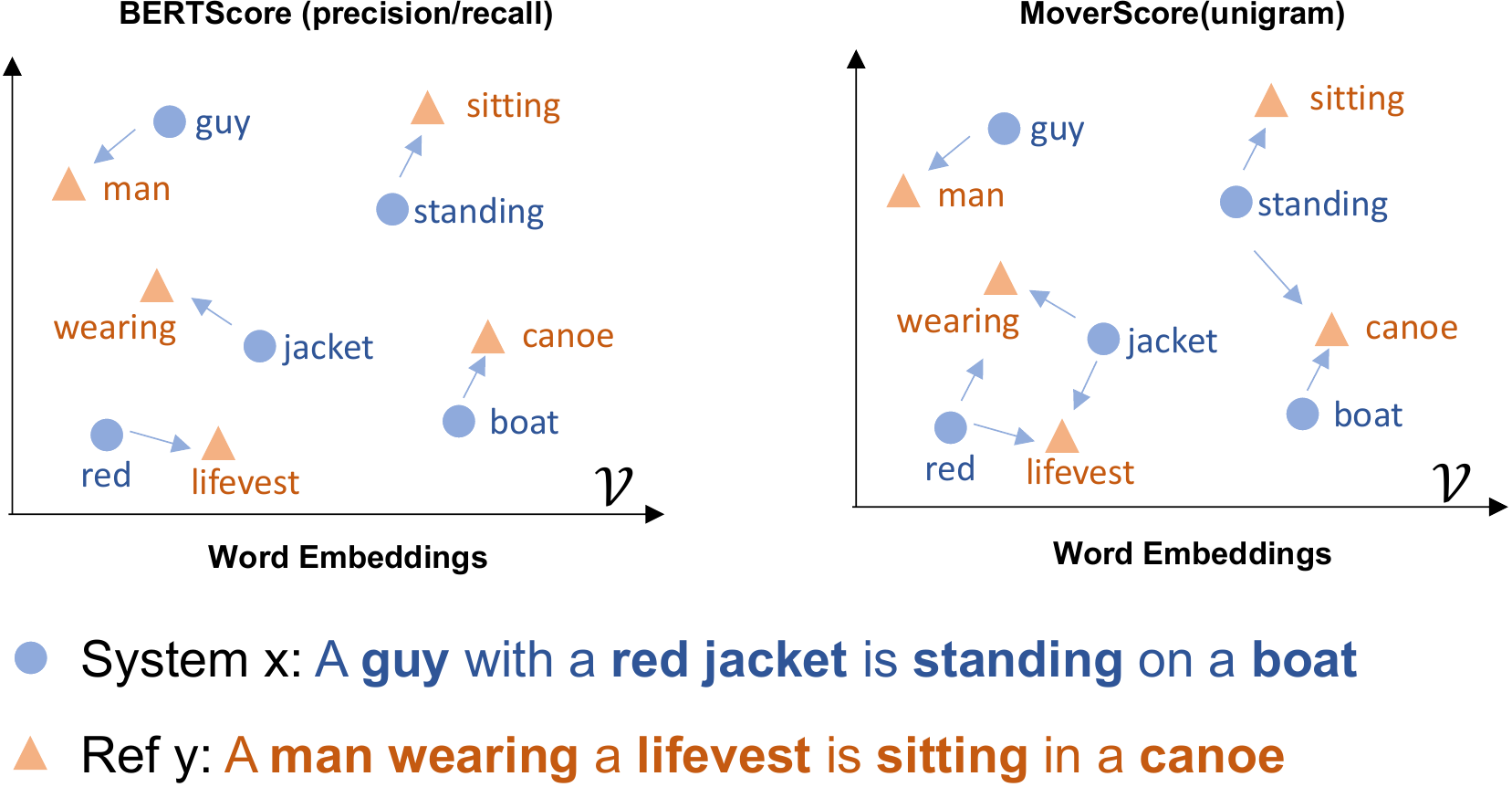}
\caption{An illustration of MoverScore and BERTScore. 
% For ease-of-read, functional words are reasonably ignored since they can be downplayed by giving them the lower weights.
}
 \label{fig:bertscore-wmd} 
\vspace{-0.2in}
\end{figure} 

% Word Mover Distance %, which 
% solves a constrained optimization problem %and the score indicates 
% to find the minimum amount of effort required to transform between two texts via a soft distance measure based on embeddings.\footnote{The formulation also provides an important possibility to bias the metric towards precision or recall by using an asymmetric cost matrix. We will discuss this later.} \todo{SE: now WMD sounds like soft edit distance. Maybe move this text up}
% Any deviation of the system text from its reference, i.e., containing more or less content, can lead to an increased distance score. 

% In addition, Mover Distance is an alternatively symmetric solution compared to asymmetric BERTScore (precision/recall).
% Mathematically, BERTScore (precision/recall) is asymmetric for a pair sentence, whereas F1 is symmetric by taking the harmonic average of the precision and recall. Instead, Mover Distance is symmetric since ... 
% For %better 
% empirical comparison of them, %with them, 
% we %disentangle 
% write BERTScore as the harmonic mean of BERTScore-Precision and BERTScore-Recall, where both two can be 
% decomposed 
% %BERTScore into two factors: 
% as a combination of ``Hard Mover Distance'' (HMD) and BERT: 
%We first study the relation between MoverScore and BERTScore %mathematically:
\begin{prop}\label{prop:main}
% The precision and recall of BERTScore are the hard forms of Word Mover Distance, and BERTScore F1 is the harmonic-mean of the above two forms.
BERTScore (precision/recall) can be represented as a (non-optimized) Mover Distance 
$\langle \mC, \mF \rangle$,
% $\sum C\cdot F$, 
where $\mC$ is a transportation cost matrix based on BERT and $\mF$ is a %non-optimized 
uniform 
transportation flow matrix.\footnote{See the proof in the appendix.}
\end{prop}
% \todo{SE: below the notation was $(C,F)$ (but in angle brackets and using bold font), now the notation is $\sum C\cdot F$. Is there a reason for this? Or is it just a nice way to confuse the reader?}

\subsection{Contextualized Representations}
% \subsection{Aggregated Contextualizers}

% Now, we propose another way to leverage contextualized word vectors produced techniques like ELMO and BERT.

The task formulation naturally lends itself to deep contextualized representations for inducing word vectors $E(\ervx_i)$.
%\todo{SE: see above. Why not bold? Wei: it's corrected to represent one word vector, bold notation means n-gram vectors}
Despite the recent success of multi-layer attentive neural architectures~\cite{Devlin:2018,Peters:2018}, consolidating layer-wise information remains an open problem
% is.\todo{SE: However, despite?}
as different layers capture information at disparate scales and task-specific layer selection methods may be limited~\cite{Liu:2018:EMNLP,liu:2019}. \citet{tenney:2019} found that %using 
a scalar mix of output layers trained from task-dependent supervisions would be effective in a deep transformer-based model. Instead, we investigate aggregation functions to consolidate layer-wise information, forming stationary representations of words without 
%the help of 
supervision.

% Recently, \citet{liu:2019} observed that the intermediate layers of contextualized representations produce strong performance, whereas the initial layers are usually underrepresented\todo{SE: they underperform? FL: I clarified the text a bit.}. In general, however, it is not clear which layer has best performance and it is most likely highly task-dependent. In order to produce robust metrics that generalize across NLG tasks, we investigate various techniques to aggregate the word representations  given by each layer into one `stronger' vector.

% Which layer has the best performance is highly task-dependent. One could decide this question for each NLG evaluation task by comparing each layer representation on its ability to correlate with human judgments for the specific task, but
% Meanwhile, which layer leads to the best performance is a mystery and highly task-dependent. While one can simply find the best layer in contextualzers with human provided judgments, it is impractical because one cannot generalize it to a new dataset without human scores. Instead of a lottery-choice from the intermediate layers, we consider aggregating contexualized word vectors coming from different layers into one stronger and richer vector, expecting to reach to the oracle performance obtained from best layers.

% especially the performance of
% Thus, we consider aggregating the vectors coming from the different layers into one stronger and richer vector.

Consider a sentence $\bx$ passed through contextualized encoders such as ELMo and BERT with $L$ layers. Each layer of the encoders produces a vector representation for each word $\ervx_i$ in $\bx$.
%\todo{SE: strictly, a sentence is not a set...} 
We denote by  $\bz_{i,l} \in \mathbb{R}^d$ the representation given by layer $l$, a $d$-dimensional vector. %Finally, 
Overall, 
$\ervx_i$ receives $L$ different vectors $(\bz_{i,1}, \dots, \bz_{i,L})$.
%\todo{SE: remember the definition of a set, which excludes copies. I think you want a tuple here, not a set. Tuple has round brackets. Wei: thanks. updated.here's a sequence} 
An aggregation $\phi$ maps these $L$ vectors to one final vector:
\begin{equation*}
E(\ervx_i) = \phi(\bz_{i,1}, \dots, \bz_{i,L})
\numberthis
\end{equation*}
where $E(\ervx_i)$ is the aggregated representation of the word $\ervx_i$.
% \todo{SE: this explains how unigrams are treated. How are bi- and trigrams treated? I saw the answer, it's in Eq. (1), but maybe it's good to help the reader here?}
%\todo{SE: I wrote in $R^d$. But I assume it could also be a sub- or superspace.}

% Let $T = |S_x^n|$ be the number of $n$-grams in the sentence $S_x^n$.
% Then, each $n$-gram $\bv_{x_i} \in S_x^n$ receives $L$ different vectors $(\mathbf{z}_{i,1}, \dots, \mathbf{z}_{i,L})$, where $d$ is the dimensionality of the vector representations and $L$ is the number of layers in the architecture considered.

% where $\mathbf{v}_i$ is the aggregated vector representing $\bv_{x_i}$ obtained from the $L$ vectors $\mathbf{z}_i \in \mathbb{R}^d$.
% When applied to each $n$-gram, it gives the final representation of the sentence $S_x^n$.

%A simple choice for $\phi$ %consists of %simply 
%concatenates %ing 
%each representation, resulting in a $d \cdot L$-dimensional vector for %$E(\ervx_i)$. However, in our experiments, concatenation worked %poorly.\todo{SE: if we need space, remove it. It is not very surprising} % Therefore, 
We study two alternatives for $\phi$: %like 
(i) the concatenation of power means \cite{Rckl2018ConcatenatedPW} as a generalized pooling mechanism, and 
%, which narrow the gap to more sophisticated sentence embeddings such as InferSent \cite{ConneauKSBB17}. 
%We also explored another principled aggregation function with 
(ii) a routing mechanism for aggregation~\cite{zhao:2018,zhao-etal-2019-towards}.
% \cite{Sabour:2017}.
We relegate the routing method to appendix, as it does not yield better results than power means.
% and consumes more space to describe.
%\todo{FL: I wasn't sure about the term "relegate" and if there's a more proper one to use. SE: it's common in economics -- "relegate to the appendix" has many hits}

% Recently, aggregation of word embeddings using concatenations of power means \cite{Rckl2018ConcatenatedPW} (as generalized pooling mechanisms) %demonstrated 
%showed improvements by concatenating power means
%of the embeddings, closing 

\paragraph{Power Means}
Power means 
% \todo{SE: we can think about whether we should really introduce the power mean notation, as we are effectively not using it. Reviewers may dislike it. We only ever use min/max/average}{} 
%also known as generalized mean, 
is an effective %powerful %substitute for 
generalization of pooling techniques for aggregating information. 
It computes a non-linear average of a set of values with an exponent $p$ (Eq.~(\ref{eq:power_means})).
%A small $p$ shifts the average towards the min value and a large $p$ emphasizes the max value.
Following \citet{Rckl2018ConcatenatedPW}, we exploit power means to aggregate vector representations $(\bz_{i,l})_{l=1}^L$ pertaining to the $i$-th word from all layers of a deep neural architecture.
Let $p\in\mathbb{R}\cup\{\pm\infty\}$, the $p$-mean of $(\bz_{i,1}, \dots, \bz_{i,L})$ is: %given by:
\begin{align*}
  %\mathbf{h}_j^p =
  \bh_i^{(p)}=\left(\frac{\bz_{i,1}^p+\cdots+\bz_{i,L}^p}{L}\right)^{1/p} \in \R^d%;\quad p\in\mathbb{R}\cup\{\pm\infty\}
  \label{eq:power_means}
  \numberthis
\end{align*}
where %addition and  \todo{SE: + is naturally defined for vectors, exponentiation isn't. I dropped it to save space}
exponentiation is applied elementwise. %to vectors. 
This generalized form can induce common named means such as arithmetic mean ($p=1$) and geometric mean ($p=0$). In extreme cases, a power mean reduces to the minimum value of the set when $p=-\infty$, and the maximum value when $p=+\infty$. The concatenation of $p$-mean vectors we use in this paper is denoted by:
\begin{align*}
  E(x_i) = %\mathbf{h}_j^{p_1}\oplus\cdots\oplus \mathbf{h}_j^{p_K}.
  \mathbf{h}_i^{(p_1)}\oplus\cdots\oplus \mathbf{h}_i^{(p_K)}
  \numberthis
\end{align*}
where $\oplus$ 
% represents concatenation of vectors.
is vector concatenation;
$\{p_1,\ldots,p_K\}$ are exponent values, and we use $K=3$ with 
%$p \in \{1, \pm\infty\}$ 
$p=1,\pm\infty$ 
in this work. 
% Fei's note: space allows perhaps the below sentence can be put in a footnote
% We further explored $p=2,3$ but did not observe improvements. 
%\todo{SE: there was a sub-index $j$, but I don't know what it stood for}

% \todo[inline]{SE: I assume the static embeddings need no further discussion?}

% A simple aggregation %consists in 
% is the component-wise average of $\bz_{i,l}, l \in \{1, \dots, L\}$.
% \citet{Rckl2018ConcatenatedPW} observed that the average is only one summary of %the 
% sequences of embeddings and other averages ($p$-means) could capture complementary information. They propose to concatenate multiple $p$-means in order to produce a final aggregate representation of a sequence of vectors.

% In this paper, we
% employ power means to output aggregated representations from contextualizers,\todo{SE: unclear. Is "output" a noun or a verb here? how about: "we aggregate representations from contextualizers using power means, defined as"}  defined as:

\subsection{Summary of MoverScore Variations}
% \todo{FL: If space is a concern I feel this section perhaps could be shortened. It may also be moved to the beginning of Section 4 (after describing the "Sentence Mover.") and this change is optional.}
%The 
%Variation in our approach comes in four different kinds: 
% Our approach %varies across four different dimensions: 
% has four dimensions: 
We investigate our MoverScore along four dimensions:
(i) the granularity of embeddings, i.e., the size of $n$ for $n$-grams, (ii) the choice of pretrained embedding mechanism, (iii) the fine-tuning task used for BERT\footnote{ELMo usually requires heavy layers on the top, which restricts the power of fine-tuning tasks for ELMo.} (iv) the aggregation technique ($p$-means or routing) when applicable.

\paragraph{Granularity}
We used $n = 1$ and $n = 2$ as well as full sentences ($n=$ size of the sentence).

\paragraph{Embedding Mechanism}
We obtained word embeddings from three different methods: \emph{static} embedding with word2vec as well as contextualized embedding with ELMo and BERT. If $n > 1$, $n$-gram embeddings are calculated by Eq.~(\ref{eq:n-gram-embedding}). Note that they represent sentence embeddings when $n=$ size of the sentence.
% \todo{SE: specificially computed sentence embeddings, with a weighted average over BERT embeddings or Word2Vec embeddings}

\paragraph{Fine-tuning Tasks}
%\citet{Devlin:2018} demonstrated the impact of the pretraining task with respect to contextualizer performances. However, pretraining on language modeling tasks might not adequately capture the semantic meaning of sentences. In contrast, 
Natural Language Inference (NLI) and paraphrasing pose high demands in understanding sentence meaning. %semantics 
%due to the semantic nature of NLI, 
%which 
This motivated us to
% We further 
%study the transferability of more related tasks with finetuning techniques 
fine-tune BERT representations on two NLI datasets, MultiNLI and QANLI, and one Paraphrase dataset, QQP---the largest datasets in GLUE \cite{wang:2018}.
%for fune-tuning tasks comparison, %\todo{SE: Is QQP also an NLI task? Wei:fixed} 
We fine-tune BERT on each of these, yielding different contextualized embeddings for our general evaluation metrics.
%We expect enhanced embeddings, %would 
%which improve the semantic textual similarity for a sentence pair. 
%\todo{SE: pretraining or fine-tuning? Wei: fixed, but need to be shortened}
%Therefore, we would study the transferability of pretraining tasks to the evaluation of NLG systems.
%\todo{SE: how do pretraining tasks transfer to evaluation? Sounds odd. Wei: fixed}

\paragraph{Aggregation}
For ELMo, we aggregate %the 
word representations given by all %of 
three ELMo layers, using %the techniques presented above: 
$p$-means or routing (see the appendix). 
%while the 
Word representations in BERT are aggregated from the last five layers, using $p$-means or routing since the representations in the initial layers are %demonstrated 
%not transferable %in the 
less suited 
for use in  
downstream tasks \cite{liu:2019}.

\section{Empirical Evaluation}

In this section, we measure the quality of different metrics on four tasks:
% including our %Mover Distance 
% own metric, 
%variations upon two families of Mover Distance: Word Mover and Sentence Mover, 
\emph{machine translation}, \emph{text summarization}, 
\emph{image captioning} and \emph{dialogue generation}.
Our major focus is to study the correlation between
different metrics and human judgment.
We employ two text encoders to embed $n$-grams:
$\bertbase$, which uses a 12-layer transformer,
and $\elmobase$, which uses a 3-layer BiLSTM.
%We use uncased English $\bertbase$ model for all of these English tasks,  which
%uses a 12-layer transformer, while $\elmobase$ model we study in this work uses a 3-layer biLSTM.
We use Pearson's $r$ and Spearman's $\rho$ to measure the correlation.
%Pearson's $r$ reflects the noisiness of a linear 
%relationship between two variables. 
%Spearman's $\rho$ is the Pearson correlation between their rank values.
%
%The notations of Mover Distance are denoted as:
We consider two variants of MoverScore: \emph{word mover} 
and \emph{sentence mover}, described below. 
%with the following notations:

\vspace{0.05in}
\noindent\textbf{Word Mover}\quad
% Our word mover metrics are denoted as
We denote our word mover notation containing four ingredients as: \emph{WMD-{Granularity}+{Embedding}+{Finetune}+{Aggregation}}. For example, $\metric{WMD}$-$1$+BERT+MNLI+PMEANS represents the semantic metric using word mover distance where unigram-based word embeddings fine-tuned on MNLI are aggregated by $p$-means.
% the metric 
% based on unigrams where the word embeddings use $p$-means to aggregate the representations given by BERT layers pretrained on MNLI.

\vspace{0.05in}
\noindent\textbf{Sentence Mover}\quad
% We denote our sentence mover variants as:
We denote our sentence mover notation with three ingredients as:
\emph{SMD+{Embedding}+{Finetune}+{Aggregation}}. For example, $\metric{SMD}$+W2V represents the semantic metric using sentence mover distance where two sentence embeddings are computed as the weighted sum over their word2vec embeddings by Eq.~(\ref{eq:n-gram-embedding}).
\setlength{\tabcolsep}{4.5pt}
\begin{table*}[h!]
    \setlength{\tabcolsep}{4pt}
    \footnotesize    
    \centering
    \begin{tabular}{l | l ccccccc c}
    \toprule
    & & \multicolumn{8}{c}{\textbf{Direct Assessment}}\\
    Setting & Metrics & cs-en & de-en & fi-en & lv-en & ru-en & tr-en & zh-en & Average \\
    % \midrule
    % \multicolumn{9}{l}{\textit{Learning metrics trained from human judgments}}\\
    \midrule
    \multirow{3}{*}{\metric{Baselines}}
    % \metric{BLEND} & 0.594 & 0.571 & 0.733 & 0.594 & 0.622 & 0.671 & 0.661 & 0.635 \\
    &\metric{METEOR++} & 0.552 & 0.538 & 0.720 & 0.563 & 0.627 & 0.626 & 0.646 & 0.610 \\
    % &\metric{chrF++} & 0.523 & 0.534 & 0.678 & 0.520 & 0.588 & 0.614 & 0.593 & 0.579 \\
    &\metric{RUSE(*)} & 0.624 & 0.644 & 0.750 & 0.697 & 0.673 & 0.716 & 0.691 & 0.685 \\
    &\metric{BERTScore-F1} & 0.670 & 0.686 & 0.820 & 0.710 & 0.729 & 0.714 & 0.704 & 0.719 \\
    % \midrule
    % \multicolumn{9}{l}{\textit{$n$-gram matching metrics}}\\
    % \midrule
    % \multirow{4}{*}{Unsupervised} 
    % \metric{BLEU-4} & 0.330 & 0.367 & 0.492 & 0.321 & 0.348 & 0.462 & 0.459 & 0.397\\
    % \metric{SentBLEU} & 0.435 & 0.432 & 0.571 & 0.393 & 0.484 & 0.538 & 0.512 & 0.481 \\
    % \metric{METEOR++} & \textbf{0.552} & \textbf{0.538} & \textbf{0.720} & \textbf{0.563} & \textbf{0.627} & \textbf{0.626} & \textbf{0.646} & \textbf{0.610} \\
    \midrule
    % \multicolumn{9}{l}{\textit{Semantic metrics via word embedding}}\\
    % \midrule
        \multirow{4}{*}{\metric{Sent-Mover}} 
    % \metric{InferSent} & 0.504 & 0.539 & 0.651 & 0.527 & 0.542 & 0.562 & 0.554 & 0.554 \\
    &\metric{\metric{Smd} + W2V} & 0.438 & 0.505 & 0.540 & 0.442 & 0.514 & 0.456 & 0.494 & 0.484 \\
    &\metric{\metric{Smd} + ELMO  + PMeans} & 0.569 & 0.558 & 0.732 & 0.525 & 0.581 & 0.620 & 0.584 & 0.595 \\
    % \metric{ELMO-P-Means} & 0.509 & 0.515 & 0.632 & 0.458 & 0.544 & 0.484 & 0.535 & 0.525 \\ P-means over T outputs bad results
    &\metric{\metric{Smd} + BERT  + PMeans} & 0.607 & 0.623 & 0.770 & 0.639 & 0.667 & 0.641 & 0.619 & 0.652 \\
    
    &\metric{\metric{Smd} + BERT  + MNLI + PMeans} & 0.616 & 0.643 & 0.785 & 0.660 & 0.664 & 0.668 & 0.633 & 0.667 \\
    % \metric{BERT-Pooling} & 0.548 & 0.577 & 0.713 & 0.580 & 0.625 & 0.572 & 0.558 & 0.596 \\
    % \midrule
    % \multirow{3}{*}{Pretraining-MNLI} 
    % \metric{BERT-IDF-Avg}& 0.616 & 0.643 & 0.785 & 0.660 & 0.664 & 0.668 & 0.633 & \textbf{0.667}   \\
    % \metric{BERT-P-Mean} & 0.564 & 0.599 & 0.727 & 0.600 & 0.624 & 0.604 & 0.578 & 0.613 \\
    \midrule
    
    \multirow{5}{*}{\metric{Word-Mover}}
    &\metric{Wmd-1 + W2V} &0.392 & 0.463 & 0.558 & 0.463 & 0.456 & 0.485 & 0.481 & 0.471 \\
    &\metric{Wmd-1 + ELMO + PMeans} & 0.579 & 0.588 & 0.753 & 0.559 & 0.617 & 0.679 & 0.645 & 0.631\\ 
    % \metric{Wmd-1 + ELMO} & 0.540 & 0.556 & 0.734 & 0.524 & 0.575 & 0.644 & 0.630 & 0.600 \\
    % \midrule
    &\metric{Wmd-1 + BERT + PMeans} & 0.662 & 0.687 & 0.823 & 0.714 & 0.735 & 0.734 & 0.719 & 0.725\\ 
    % &\metric{Wmd-1 + BERT + Best Layer} & 0.669 & 0.695 & 0.824 & 0.719 & 0.736 & 0.735 & 0.720 & 0.728\\ 
    
    % & 0.686 & 0.823 & 0.710 & 0.729 & 0.729 & 0.720 & 0.721 \\ 
    % &\metric{Wmd-2 + BERT + Best Layer} & 0.679 & 0.697 & 0.820 & 0.721 & 0.739 & 0.737 & 0.724 & 0.731 \\ 
    % & 0.688 & 0.821 & 0.712 & 0.728 & 0.735 & 0.719 & 0.724 \\
    &\metric{Wmd-1 + BERT + MNLI + PMeans} & 0.670 & 0.708 & \textbf{0.835} & \textbf{0.746} & \textbf{0.738} & 0.762 & \textbf{0.744} & \textbf{0.743}\\ 
    
    &\metric{Wmd-2 + BERT + MNLI + PMeans} & \textbf{0.679} & \textbf{0.710} & 0.832 & 0.745 & 0.736 & \textbf{0.763} & 0.740 & \textbf{0.743}\\ 

	\bottomrule  
    \end{tabular}
    % }
    \caption{Absolute Pearson correlations with segment-level human judgments in 7 language pairs on WMT17 dataset. 
    % The correlations of \metric{RUSE}, \metric{chrF++} and \metric{BERTScore-F1} are cited from [] and [].
    \label{tab:wmt17-to-en} }
\vspace{-0.1in}
\end{table*}

\paragraph{Baselines}
% We compare %two families of Mover Distance 
% against  \emph{ChrF++}\cite{popovic:2017} 
We select multiple strong baselines for each task for comparison: %metrics for 
SentBLEU, METEOR++ \cite{guo:2018}, and a supervised metric RUSE for machine translation; ROUGE-1 and ROUGE-2 and a supervised metric $S^{3}_{best}$ \cite{Peyrard:2017} for text summarization; BLEU and METEOR for dialogue response generation, CIDEr, SPICE, METEOR and a supervised metric LEIC \citep{cui:2018} for image captioning. We also report %a recently proposed 
BERTScore \cite{zhang:2019} for all tasks (see \S\ref{sec:related}). Due to the page limit, we only compare with the strongest baselines, the rest can be found in the appendix.
% , BLEND and  
% (see \S\ref{sec:related}),
% and also recently proposed %semantic metrics 
% \emph{BERTScore} \cite{zhang:2019}, together with a supervised metric \emph{RUSE} \cite{Shimanaka:2018}.

% Due to the page limit, we only %represent the 
% give the 
% performance of the %semantic
% %metrics in our paper; 
% strongest baselines (BERTScore, ChrF++, RUSE); 
% %the correlations
% %between the lexical metrics and human judgement
% the remainder 
% can be found in the Appendix.  

% \paragraph{Baselines}
% %Similarly, 
% We compare %two families of Mover Distance 
% against multiple %strong 
% baselines 
% %metrics 
% for text summarization, including \emph{ROUGE-1} and \emph{ROUGE-2} and %recent semantic metrics 
% \emph{BERTScore}, together with a supervised metric \emph{$S^{3}_{best}$}\cite{Peyrard:2017}. The performance of other lexical and semantic metrics can be found in Appendix.

\subsection{Machine Translation}
\label{sec:mt_results}

\paragraph{Data}
We obtain the source language sentences, their system 
and reference translations from the WMT 2017
news translation shared task~\cite{Bojar:2017}.
We consider 7 language pairs: from German (de), Chinese (zh), 
Czech (cs), Latvian (lv), Finnish (fi), Russian (ru), 
and Turkish (tr), resp.\ to English. 
Each language pair has approximately 3,000 sentences,
and each sentence has one reference translation and multiple
system translations generated by participating systems. 
For each system translation, at least 15 human 
% evaluators 
assessments
are independently rated %upon its 
for quality. 

\paragraph{Results}
Table \ref{tab:wmt17-to-en}: 
%compares Pearson's 
%$r$ %correlations 
%between different metrics and segment(sentence)-level human judgments.
%%%%%correlations on WMT17 to-English translations. 
%RUSE \cite{Shimanaka:2018} is previous state of the art in MT metric tasks, learned from human judgments on WMT16 and WMT15 datasets. ChrF++ \cite{popovic:2017} is a n-gram metric which compares n-grams in both word and character condition. BERTScore \cite{zhang:2019} is a recent semantic metric based on BERT. 
% We compare our metric variations with these baselines in 7 language pairs. 
% We perform ablation studies to study the importance of each component in metric
% also compare semantic metrics with $n$-gram WMD via different embeddings, such as static embedding and contextual word embeddings with best layer. 
In all language pairs, the best correlation is achieved
by our word mover metrics that use a BERT pretrained
on MNLI as the embedding generator and 
%use 
PMeans to aggregate the embeddings from different BERT layers, i.e., 
WMD-1/2+BERT+MNLI+PMeans.
Note that our unsupervised word mover metrics even outperforms
RUSE, a supervised metric.
%
%semantic 
%metrics, especially WMD-1+BERT+MNLI+PMeans, exceeding the correlation of other metrics, even the state-of-the-art supervised metric (RUSE)---by xx pp on average. We also find that our metrics combining contextual word embedding pretrained on the respective task (e.g., MNLI) consistently provides the best performance.
% word embedding pretrained by BERTRouting on MNLI dataset perfrom siginifcantly better than others, even beyond the supervised metrics BLEND and RUSE.
We also find that our word mover metrics 
% with word-level granularity 
outperforms the sentence mover. %based on sentence-level.
% our metric with word-level granularity
% outperforms the metrics based on sentence embeddings. 
We %guess\todo{SE: guessing is not a good thing} 
conjecture that %the reason is that 
important information is
lost in such a sentence representation while transforming the whole sequence of word vectors into one sentence embedding by Eq.~(\ref{eq:n-gram-embedding}).
% \todo{SE: which ones are using weighted averages? InferSent doesn't. Wei: fixed}

%by incorporating compositionality since the BERT tokenizer would split the meaning of each word into a set of sub-words.
% Although the results of them are on par, we suggest to enable bigrams in order to consider compositionality since one word often is split into sub-words by WordPiece (BERT tokenizer), which break down the meaning of each word. We leave the futher study of $n$-gram metrics in future work.

% We perform ablation studies to study the importance of each component in metric

% We also report semantic metrics with sentence embedding such as Word2vec [], ELMO-Pooling and BERT-Pooling, which are averaged from a sequence of word representations.

% We report semantic metrics with word embedding and sentence embedding contextualizers 
% 1. How and why does the choice of embedding types affect the metrics’ correlations in both clean and noise-corrupted conditions?
% 2. How does the choice of sentence- or word-level affect the correlations?
% 3. How does the metrics’ correlation vary across pretraining tasks?
% 4. How can we aggregate the information from all layers in recent contextualizers (ELMO and BERT)?
% 5. Which configuration of semantic metrics is more reliable and why?

\setlength{\tabcolsep}{3.5pt}
\begin{table*}[t]
		\small
		\centering
  		\begin{tabular}{l|l cc cc|cccc}
  		\toprule
 		&&\multicolumn{4}{c|}{TAC-2008}	&	\multicolumn{4}{c}{TAC-2009}\\
 		&&\multicolumn{2}{c}{\textbf{Responsiveness}}&\multicolumn{2}{c|}{\textbf{Pyramid}} &\multicolumn{2}{c}{\textbf{Responsiveness}} & \multicolumn{2}{c}{\textbf{Pyramid}}\\
 		Setting & Metrics	&	$r$ & $\rho$& $r$   & $\rho$ & $r$ & $\rho$ & $r$   &$\rho$\\
 		\midrule
 		
%  		\multicolumn{9}{l}{\textit{Learning metrics trained from human judgments}}\\
%         \midrule
        \multirow{4}{*}{\metric{Baselines}}
        % \metric{$S^{3}_{full}$} & 0.696 & 0.558 & 0.753 & 0.652 & 0.731 & 0.552 & 0.838 & 0.724 \\
        % \midrule
        % \multicolumn{9}{l}{\textit{$n$-gram matching metrics}}\\
        % \midrule
        % \multirow{9}{*}{Unsupervised}
% 		&\metric{TF$*$IDF-1}	& 0.176 & 0.224 & 0.183 & 0.237 & 0.187 & 0.222 & 0.242 & 0.284\\
% 		&\metric{TF$*$IDF-2}	& 0.047 & 0.154 & 0.049 & 0.182 & 0.047 & 0.167 & 0.097 & 0.233\\
        &\metric{$S^{3}_{best}$}(*) & 0.715 & 0.595 & 0.754 & 0.652 & 0.738 & \textbf{0.595} & \textbf{0.842} & \textbf{0.731} \\
		&\metric{ROUGE-1} & 0.703 & 0.578 & 0.747 & 0.632 & 0.704 & 0.565 & 0.808 & 0.692 \\
		&\metric{ROUGE-2} & 0.695 & 0.572 & 0.718 & 0.635 & 0.727 & 0.583 & 0.803 & 0.694 \\
% 		\metric{ROUGE-1-WE}	& 0.571 & 0.450 & 0.579 & 0.458 & 0.586 & 0.437 & 0.653 & 0.516 \\
% 		\metric{ROUGE-2-WE}	& 0.566 & 0.397 & 0.556 & 0.388 & 0.607 & 0.413 & 0.671 & 0.481 \\
% 		&\metric{ROUGE-L} & 0.681 & 0.520 & 0.702 & 0.568 & \textbf{0.730} & 0.563 & 0.779 & 0.652 \\
		&\metric{BERTScore-F1} & 0.724 & 0.594 & 0.750 & 0.649 & 0.739 & 0.580 & 0.823 & 0.703 \\
% 		\metric{Frame-1} & 0.658 & 0.508 & 0.686 & 0.529 & 0.678 & 0.527 & 0.762 & 0.628 \\
% 		\metric{Frame-2} & 0.676 & 0.519 & 0.691 & 0.556 & 0.715 & 0.555 & 0.781 & 0.648 \\
	\midrule
% % 		Pyramid				& .7030 & .6604 & .8528 & --- & --- & --- & 						.7152 & .6386 & .8520 & --- & --- & ---  \\
    % \multicolumn{9}{l}{\textit{Semantic metrics via word embedding}}\\
    % \midrule
    \multirow{4}{*}{\metric{Sent-Mover}} 
    &\metric{SMD + W2V} & 0.583 & 0.469 & 0.603 & 0.488 & 0.577 & 0.465 & 0.670 & 0.560 \\
    % & 0.633 & 0.507 & 0.656 & 0.545 & 0.648 & 0.479 & 0.733 & 0.568 \\
    &\metric{SMD + ELMO + PMeans} & 0.631 & 0.472 & 0.631 & 0.499 & 0.663 & 0.498 & 0.726 & 0.568 \\
    % \metric{ELMO-P-Mean} & 0.683 & 0.504 & \textbf{0.706} & \textbf{0.554} & 0.726 & 0.516 & \textbf{0.785} & \textbf{0.614} \\
    &\metric{SMD + BERT + PMeans} & 0.658 & 0.530 & 0.664 & 0.550 & 0.670 & 0.518 & 0.731 & 0.580 \\
    &\metric{SMD + BERT + MNLI + PMeans} & 0.662 & 0.525 & 0.666 & 0.552 & 0.667 & 0.506 & 0.723 & 0.563 \\
    
    % & 0.657 & 0.503 & 0.659 & 0.517 & 0.660 & 0.460 & 0.709 & 0.518\\
    % \metric{InferSent} & 0.667 & \textbf{0.546} & 0.707 & \textbf{0.598} & 0.663 & \textbf{0.528} & \textbf{0.764} & \textbf{0.640} \\
    % \metric{BERT-P-Mean} & \textbf{0.701} & \textbf{0.558} & 0.702 & 0.535 & \textbf{0.739} & \textbf{0.538} & 0.769 & 0.609 \\
    % \midrule
    % \multirow{3}{*}{Pretraining-MNLI} 
    % & \metric{BERT-IDF-Avg} & 0.657 & 0.503 & 0.659 & 0.517 & 0.660 & 0.460 & 0.709 & 0.518\\
    % & \metric{BERT-P-Mean} & \textbf{0.705} & 0.545 & \textbf{0.709} & 0.579 & \textbf{0.729} & 0.506 & 0.761 & 0.597 \\
    \midrule

    \multirow{5}{*}{\metric{Word-Mover}}
    &\metric{Wmd-1 + W2V} & 0.669 & 0.549 & 0.665 & 0.588 & 0.698 & 0.520 & 0.740 & 0.647 \\
    &\metric{Wmd-1 + ELMO + PMeans} & 0.707 & 0.554 & 0.726 & 0.601 & 0.736 & 0.553 & 0.813 & 0.672 \\
    &\metric{Wmd-1 + BERT + PMeans} & 0.729 & 0.595 & 0.755 & 0.660 & 0.742 & 0.581 & 0.825 & 0.690 \\
    &\metric{Wmd-1 + BERT + MNLI + PMeans} & \textbf{0.736} & \textbf{0.604} & \textbf{0.760} & \textbf{0.672} & \textbf{0.754} & 0.594 & 0.831 & 0.701 \\
    &\metric{Wmd-2 + BERT + MNLI + PMeans} & 0.734 & 0.601 & 0.752 & 0.663 & 0.753 & 0.586 & 0.825 & 0.694 \\
		\bottomrule  
		\end{tabular}
 		\caption{Pearson $r$ and Spearman $\rho$ correlations with summary-level human judgments on TAC 2008 and 2009. 
%  		The correlations of $S^{3}_{best}$, \metrics{ROUGE-1}, \metrics{ROUGE-2} and \metrics{ROUGE-L} are cited from [].
 		}
 		\label{tab:tac0809}
\vspace{-0.08in}
	\end{table*}

\subsection{Text Summarization}
\label{sec:summ_results}

We use two summarization datasets from the Text Analysis Conference 
(TAC)\footnote{\url{http://tac.nist.gov}}:
TAC-2008 and TAC-2009, which %\footnote{\url{http://tac.nist.gov/2009/Summarization/}, \url{http://tac.nist.gov/2008/Summarization/}}
%TAC-2008 and TAC-2009 
contain 48 and 44 \emph{clusters}, respectively. 
Each cluster includes 10 news articles (on the same topic),
four reference summaries,
and 57 (in TAC-2008) or 55 (in TAC-2009) system 
summaries generated by the participating systems.
Each summary (either reference or system) has fewer than 100 words,
and receives two human judgment scores:  
% We use only the so-called initial summaries (A summaries), but not the update part.
%In both editions, all system summaries and the 4 reference summaries were manually evaluated by NIST assessors 
the \emph{Pyramid} score~\cite{nenkova-passonneau:2004:HLTNAACL} 
and the \emph{Responsiveness} score. 
Pyramid measures how many important semantic content units %(SCUs) 
in the reference summaries are covered by the system summary,
while Responsiveness measures %on 
how well a summary responds to the overall quality combining both content and linguistic quality.
% on how well a summary 
% responds to a general information need.
%At the time of the shared tasks, 57 systems were submitted to TAC-2008 and 55 to TAC-2009.

% We use two typical multi-document summarization
% datasets so-called TAC-2008 and TAC-2009 from the Text Analysis Conference (TAC) shared tasks [], where TAC-2008 and TAC-2009 contain source articles, system summaries from various summarization systems, reference summaries and two summary-level human scores. The human scores, pyramid and overall responsiveness, were manually evaluated by NIST assessors for readability and context selection, which associate a score with each system summaries and four reference summaries. 

% \paragraph{Baselines}
% %Similarly, 
% We compare %two families of Mover Distance 
% against multiple %strong 
% baselines 
% %metrics 
% for text summarization, including \emph{ROUGE-1} and \emph{ROUGE-2} and %recent semantic metrics 
% \emph{BERTScore}, together with a supervised metric \emph{$S^{3}_{best}$}\cite{Peyrard:2017}. The performance of other lexical and semantic metrics can be found in Appendix.
% Here we report the results from $S^{3}_{best}$ \cite{Peyrard:2017}, the best performing baseline together with ROUGE-1 and ROUGE-2 variants.
% We also report the scores obtained by BERTScore. 
\paragraph{Results}
Tables \ref{tab:tac0809}: 
%compares Spearman's $\rho$ and Pearson's $r$ correlations between different metrics and summary-level human judgments.
%$S^{3}_{best}$ \cite{Peyrard:2017} is previous state-of-the-art for evaluating text summarization, learned from responsiveness and pyramid scores on TAC 2008 and 2009 datasets in leave-one-out configuration. 
We observe that lexical metrics like ROUGE correlate above-moderate on TAC 2008 and 2009 datasets. In contrast, these metrics perform poorly on other tasks like Dialogue Generation \cite{novikova-etal-2017-need} and Image Captioning \cite{AndersonFJG16}. %We speculate that 
Apparently, strict matches on surface forms seems reasonable for \emph{extractive} summarization datasets.
% since they are extractive summarization datasets, where strict matches seems reasonable.
% while they perform poorly in other tasks (e.g., Dialog Reponse Generation, Image Captioning) since 
% see $n$-gram matching metrics (e.g., ROUGE) correlates above-moderate correlation on both datasets since they are extractive summarization datasets, where strict matches seems reasonable. 
However, we still %can observe our 
see that our word mover metrics, i.e., WMD-1+BERT+MNLI+PMeans, %exceeding or rivaling 
perform better than or %on par with 
come close to 
even the %best performing 
supervised metric $S^{3}_{best}$. 
% in a clear margin on TAC 2008, and rivals on TAC 2009.

% our metric variations combining contextual word embedding perform better at system-level correlation, rivaling or exceeding the performance of the state-of-the-art supervised metric $S^{3}_{best}$. Consistent with MT, we also find that contextual word embedding pretrained on the respective task (e.g., MNLI task) leads to an improvement.
% ROUGE-n [] is a n-gram metric which compares.... As similar in MT, we compare semantic metrics with $n$-gram WMD via different embeddings. 

\setlength{\tabcolsep}{8.3pt}
\begin{table*}
	\small
	\centering
	\begin{tabular}{l|l|c c c|c c c}
		\toprule
% 		\noalign{\smallskip}
		&& \multicolumn{3}{c}{BAGEL} & \multicolumn{3}{c}{SFHOTEL} \\
        Setting&Metrics & \textbf{Inf} & \textbf{Nat} & \textbf{Qual} & \textbf{Inf} & \textbf{Nat} & \textbf{Qual} \\
		\midrule
		\noalign{\smallskip}
        % \multicolumn{6}{l}{\textit{$n$-gram matching metrics}}\\
        % \midrule
        \multirow{4}{*}{\metric{Baselines}}
        &\metric{BLEU-1} & 0.225 & 0.141 & 0.113 & 0.107 & 0.175 & 0.069 \\
        &\metric{BLEU-2} & 0.211 & 0.152 & 0.115 & 0.097 & 0.174 & 0.071 \\
        % &\metric{BLEU-3} & 0.191 & 0.150 & 0.109 & 0.089 & 0.161 & 0.070 \\
        % &\metric{BLEU-4} & 0.175 & 0.141 & 0.101 & 0.084 & 0.104 & 0.056 \\
        % &\metric{ROUGE-L} & 0.202 & \textbf{0.134} & 0.111 & 0.092 & 0.147 & 0.062 \\
        % &\metric{NIST} & 0.207 & 0.089 & 0.056 & 0.072 & 0.125 & 0.061 \\
        % &\metric{CIDEr} & 0.205 & 0.162 & \textbf{0.119} & 0.095 & 0.155 & 0.052 \\
        &\metric{METEOR} & 0.251 & 0.127 & 0.116 & 0.111 & 0.148 & 0.082 \\
        &\metric{BERTScore-F1} & 0.267 & 0.210 & \textbf{0.178} & 0.163 & 0.193 & 0.118 \\
    % \midrule
    % \multicolumn{6}{l}{\textit{Semantic metrics via word embedding}}\\
    \midrule
    
    \multirow{4}{*}{\metric{Sent-Mover}} 
    &\metric{SMD + W2V} & 0.024 & 0.074 & 0.078 & 0.022 & 0.025 & 0.011 \\
    &\metric{SMD + ELMO + PMeans} & 0.251 & 0.171 & 0.147 & 0.130 & 0.176 & 0.096 \\
    % \metric{ELMO-P-Mean} & 0.231 & 0.147 & 0.127 & 0.137 & 0.190 & 0.109 \\
    &\metric{SMD + BERT + PMeans} & 0.290 & 0.163 & 0.121 & 0.192 & 0.223 & 0.134 \\
    % \metric{BERT-P-Mean} & 0.291 & 0.147 & 0.105 & \textbf{0.208} & \textbf{0.224} & \textbf{0.140} \\
    % \midrule
    % \multirow{3}{*}{Pretraining-MNLI} 
    &\metric{SMD + BERT + MNLI + PMeans} & 0.280 & 0.149 & 0.120 & 0.205 & 0.239 & 0.147 \\
    % \metric{BERT-P-Mean} & 0.277 & 0.136 & 0.103 & \textbf{0.227} & 0.237 & \textbf{0.151} \\
    % & \metric{InferSent} & 0.254 & 0.096 & 0.096 & 0.157 & 0.132 & 0.052 \\
        
        % \metric{BERT-Pooling} &\textbf{0.269} & \textbf{0.182} & \textbf{0.146} & \textbf{0.238} & \textbf{0.219} & \textbf{0.117}\\
        % \metric{BERT-Routing} & - & - & - & - & -  \\
        % \metric{Infersent} & 0.202 & 0.081 & 0.052 & 0.178 & 0.146 & 0.038  \\

    \midrule
    
    \multirow{5}{*}{\metric{Word-Mover}}
    &\metric{Wmd-1 + W2V} & 0.222 & 0.079 & 0.123 & 0.074 & 0.095 & 0.021 \\
    &\metric{Wmd-1 + ELMO + PMeans} & 0.261 & 0.163 & 0.148 & 0.147 & 0.215 & 0.136  \\
    &\metric{Wmd-1 + BERT + PMeans} & \textbf{0.298} & \textbf{0.212} & 0.163 & 0.203 & 0.261 & 0.182  \\
    &\metric{Wmd-1 + BERT + MNLI + PMeans} & 0.285 & 0.195 & 0.158 & \textbf{0.207} & \textbf{0.270} & \textbf{0.183}  \\
    &\metric{Wmd-2 + BERT + MNLI + PMeans} & 0.284 & 0.194 & 0.156 & 0.204 & 0.270 & 0.182  \\
		\bottomrule  
	\end{tabular}
	\caption{Spearman correlation with utterance-level human judgments for BAGEL and SFHOTEL datasets. 
% 	The correlation of \metric{ROUGE-L} and \metric{METEOR} are cited from [].
	}
	\label{tab:bag_sfh}
\vspace{-0.1in}
\end{table*}
\setlength{\tabcolsep}{1.4pt}
\subsection{Data-to-text Generation}
\label{sec:gen_results}
We use two task-oriented dialogue datasets: BAGEL \cite{mairesse:2010} and SFHOTEL \cite{wen:2015}, which contains 202 and 398 instances of Meaning Representation (MR). Each MR instance includes multiple references, and roughly two system utterances generated by different neural systems. Each system utterance receives three human judgment scores: \emph{informativeness}, \emph{naturalness} and \emph{quality} score \cite{novikova-etal-2017-need}. Informativeness %concerns 
measures 
how much information a system utterance %is provided against 
provides with respect to an MR. Naturalness measures how likely a system utterance is generated by native speakers. Quality %
measures 
%concerns 
how well a system utterance 
%responds to the grammatical correctness and fluency. 
captures fluency and grammar. 
\paragraph{Results}
% both of which aim to generate utterance from semantic frames, where BAGEL and SFHOTEL provide the information of restaurants and hotels in Cambridge and  San Francisco.
% provides restaurants information in Cambridge while SFHOTEL provides hotels information in San Francisco. 
% and both of them contain semantic frames and corresponding references. 
% Each utterance reviews three human judgement scores: \emph{informativeness}, \emph{naturalness} and \emph{quality} score \citet{novikova-etal-2017-need}.

% We obtained system utterances and utterance-level human judgments (informativeness, naturalness and quality) from .
% The human scores are evaluated upon informativeness, naturalness and quality.
% associating a score with each utterance and a couple of references. 
Tables \ref{tab:bag_sfh}:
%compares Spearman correlations between different metrics and utterance-level human judgments. 
Interestingly, 
%we observe that 
no metric produces an even moderate correlation with human judgments, including our own. We speculate that current contextualizers are poor at representing named entities like hotels and place names as well as numbers appearing in system and reference texts. 
%\todo{SE: static embeddings also? Wei: updated. SE: Don't see it. It still only speaks about how bad contextualizers are --- are the static embeddings better?} 
However, %we still can observe that 
%the 
best correlation is still achieved by our word mover metrics combining contextualized representations.
% , no metric produces an even moderate
% correlation with human ratings
% We observe that  
% BLEU-n [] is a n-gram metric which compares.... As similar in MT, we compare semantic metrics with $n$-gram WMD via different embeddings. 

\subsection{Image Captioning}
\label{sec:cap_results}
We use a %well-known 
popular 
image captioning dataset: MS-COCO \cite{lin:2014:coco}, which contains 5,000 images. Each image includes roughly five reference captions, and 12 system captions generated by the participating systems from 2015 COCO Captioning Challenge. For the system-level human correlation, each system receives five human judgment scores: M1, M2, M3, M4, M5 \citep{AndersonFJG16}. The M1 and M2 scores %concerns the 
measure overall quality of the captions while M3, M4 and M5 scores 
%were collected for measuring the 
measure correctness, detailedness and saliency of the captions.
% from MSCOCO Captioning Challenge \footnote{ \scriptsize\url{http://cocodataset.org/#captions-leaderboard}}
% , where it provides images, corresponding human captions and five system-level human judgments that associate each system with one score in the test set. 
% The M1 and M2 scores were manually evaluated upon the overall quality of the captions while M3, M4 and M5 scores were collected for measuring the correctness, detailedness and saliency of the captions. 
Following \citet{cui:2018}, we compare the Pearson correlation with two system-level scores: M1 and M2, since we focus on studying metrics 
%upon 
for the overall quality of the captions, leaving metrics understanding captions in different aspects (correctness, detailedness and saliency) to future work.

\paragraph{Results} 
Table \ref{tab:caption}: %compares Pearson's $r$ 
%%correlation 
%between different metrics and system-level human judgments. %We observe 
% Best correlation comes from 
%\todo{YG: I would simply say `Word-Mover based metrics outperform all baselines except ...'. `Best' and `outperform all' are redundant. Wei: updated}
% Word-Mover based metrics, %which 
% outperforming all
% Our Word-Mover based metric WMD-1+BERT+MNLI+PMeans exceeds all 
%strong
% baselines except for the supervised metric LEIC, which  uses more information by considering \emph{both} images and texts. 
Word mover metrics outperform all baselines except for the supervised metric LEIC, which  uses more information by considering \emph{both} images and texts. 
%However, accessing to the images make such metric difficult to generalize to other tasks, where only text are provided.\todo{SE: don't understand Wei: fixed}
\setlength{\tabcolsep}{3.7pt}
\begin{table}[t!]
    \small
    \centering
    \begin{tabular}{l|l | c c}
    \toprule
    Setting & Metric & M1 & M2 \\
    \midrule
    % \multicolumn{3}{l}{\textit{Learning metrics trained from human judgments}}\\
    % \midrule
    \multirow{5}{*}{\textsc{Baselines}}
    &\metric{LEIC(*)} & \textbf{0.939} & \textbf{0.949} \\
    &\metric{METEOR} & 0.606 & 0.594 \\
    % &\metric{CIDEr} & 0.438 & 0.440 \\
    &\metric{SPICE} & 0.759 & 0.750 \\
    &\metric{BERTScore-Recall} & 0.809 & 0.749\\
    % \midrule
    % \multicolumn{3}{l}{\textit{$n$-gram matching metrics}}\\
    % \midrule
    % \metric{BLEU-1} & 0.124 & 0.135 \\
    % \metric{BLEU-2} & 0.037 & 0.048 \\
    
    % & \rouge-L & 0.090 & 0.096 \\
    % \midrule
    % \multicolumn{3}{l}{\textit{Semantic Frame metrics}}\\
    % \midrule
    
    \midrule
    \multirow{4}{*}{\metric{Sent-Mover}}
    &\metric{SMD + W2V} & 0.683 & 0.668\\
    &\metric{SMD + ELMO + P} & 0.709 & 0.712\\
    &\metric{SMD + BERT + P} & 0.723 & 0.747\\
    &\metric{SMD + BERT + M + P} & 0.789 & 0.784\\
    \midrule
    
    \multirow{5}{*}{\metric{Word-Mover}}
    % \multicolumn{3}{l}{\textit{Semantic metrics via word embedding}}\\
    &\metric{Wmd-1 + W2V} & 0.728 & 0.764\\
    &\metric{Wmd-1 + ELMO + P} & 0.753 & 0.775\\
    &\metric{Wmd-1 + BERT + P} & 0.780 & 0.790\\
    &\metric{Wmd-1 + BERT + M + P} & \textbf{0.813} & \textbf{0.810}\\
    &\metric{Wmd-2 + BERT + M + P} & 0.812 & 0.808\\
    % \midrule
    
    \bottomrule
    \end{tabular}
    \caption{Pearson correlation with system-level human judgments on MSCOCO dataset. 'M' and 'P' are short names.}
    \label{tab:caption}
\vspace{-0.05in}
\end{table}

\setlength{\tabcolsep}{5pt}
\begin{table}[h!]
    \footnotesize    
    \centering
    \begin{tabular}{l|cccc}
    \toprule
    % & \multicolumn{4}{c}{\textbf{Direct Assessment}}\\
    Metrics & cs-en & de-en & fi-en & lv-en \\
    % \midrule
    % \multicolumn{10}{l}{\textit{Learning Metrics trained from human judgments}}\\
    \midrule
    % \multirow{1}{*}{Existing Metrics}
    \metric{RUSE} & 0.624 & 0.644 & 0.750 & 0.697 \\
    \midrule
    % \multirow{3}{*}{Word-WMD}
    \metric{\metric{Hmd-F1} + BERT} & 0.655 & 0.681 & 0.821 & 0.712 \\
    \metric{\metric{Hmd-Recall} + BERT} & 0.651 & 0.658 & 0.788 & 0.681\\
    \metric{\metric{Hmd-Prec} + BERT} & 0.624 & 0.669 & 0.817 & 0.707 \\
    \midrule
    \metric{Wmd-unigram + BERT} & 0.651 & 0.686 & \textbf{0.823} & 0.710 \\ 
    \metric{Wmd-bigram + BERT} & \textbf{0.665} & \textbf{0.688} & 0.821 & \textbf{0.712} \\ 
    
    \hline
    \end{tabular}
    % }
    \caption{Comparison on hard and soft alignments.}
    \label{tab:wmt17-ablation}
\vspace{-0.05in}
\end{table}

\subsection{Further Analysis}
\noindent\textbf{Hard and Soft Alignments}\quad 
%Firstly, we %disentangle 
%write 
BERTScore is the harmonic mean of BERTScore-Precision and BERTScore-Recall, where both two can be 
decomposed 
%BERTScore into two factors: 
as a combination of ``Hard Mover Distance'' (HMD) and BERT (see Prop.~\ref{prop:main}).
We use the representations in the 9-th BERT layer for fair comparison of %both approaches 
BERTScore and MoverScore 
% Word-Mover
%\todo{SE: between what? BERTScore and MoverScore? Here and below it is always called Word-Mover?!?! Wei: updated}
and show results on the machine translation task in Table \ref{tab:wmt17-ablation}. %We find that BERTScore combines two asymmetric information sources (precision and recall), each of which are worse than Mover Distance. 
%We find that 
% Word-Mover, one variant of our MoverScore featured on soft aligments, 
MoverScore outperforms both asymmetric HMD factors, while if they are combined via harmonic mean, BERTScore is on par with MoverScore. We conjecture that BERT softens hard alignments of BERTScore as contextualized embeddings encode information about the whole sentence into each word vector.
%This shows that the `soft transportation' of WMD outperform the `hard transportation' of 
We also observe that \metric{Wmd-bigrams} slightly %marginally %improves the correlation 
outperforms \metric{Wmd-unigrams} on 3 out of 4 language pairs.

\begin{figure}
\begin{minipage}{0.49\linewidth}  
	\centerline{\includegraphics[width=\linewidth]{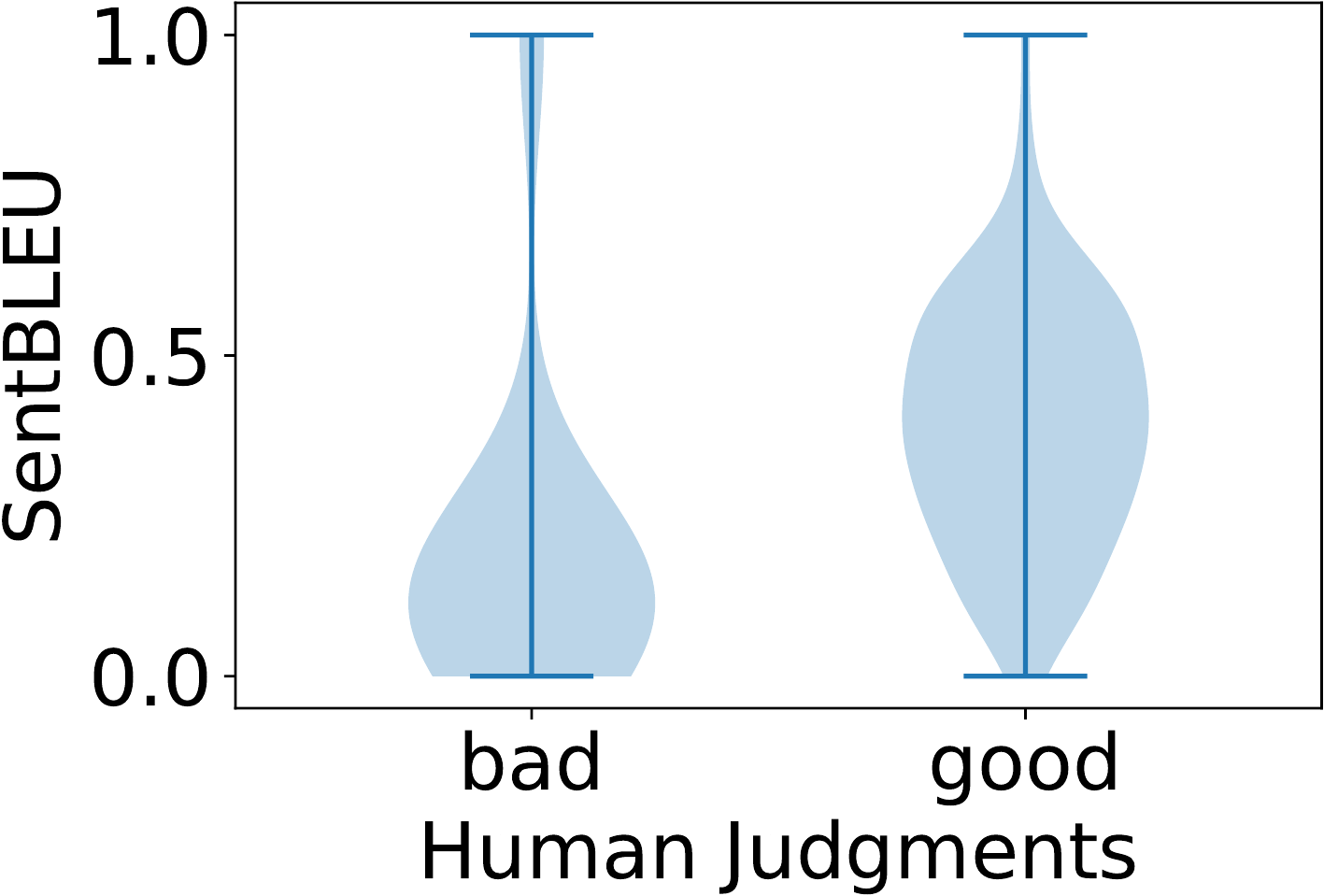}}  
% 	\centerline{(a) this is a placeholder ($n$-gram matching metrics)}  
\end{minipage} 
\begin{minipage}{0.49\linewidth}  
	\centerline{\includegraphics[width=\linewidth]{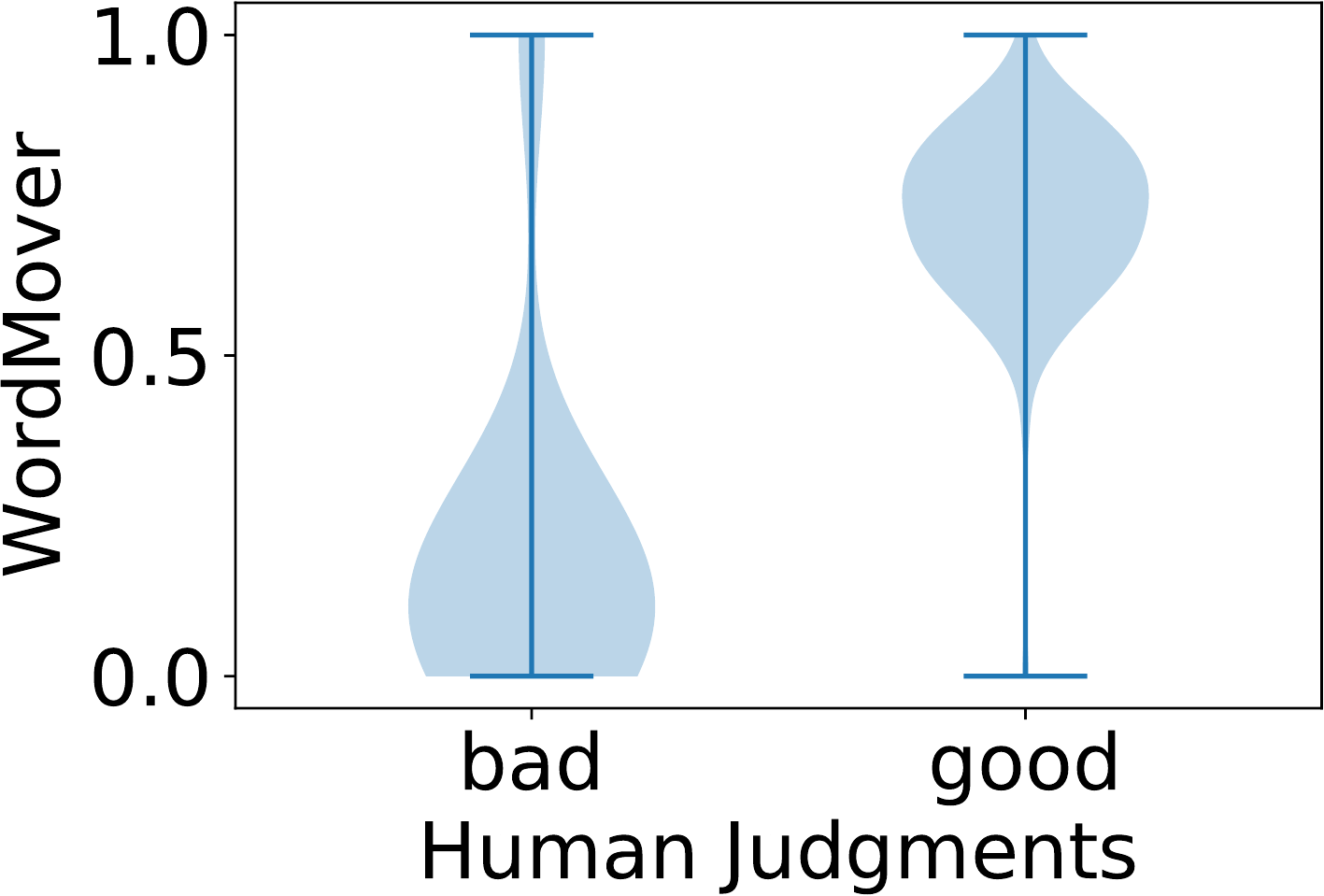}}  
% 	\centerline{(b) this is a placeholder (semantic metrics)}  
\end{minipage} 
 \caption{
 Score distribution in German-to-English pair.
%  Score distribution for one lexical matching metric \metric{SentBLEU} and one semantic metric \metric{Wmd-1-BERT} On WMT17 German-to-English dataset. \metric{SentBLEU} correctly assign low scores to low quality translations, while it struggles in judging high quality ones. Whereas, \metric{Wmd-1-BERT} can clearly distinguish both cases.
 }
 \label{fig:score_dist}  
%\vspace{-10pt}
\vspace{-0.1in}
\end{figure}

\vspace{0.1in}
\noindent\textbf{Distribution of Scores}\quad
In Figure \ref{fig:score_dist}, we take a closer look at sentence-level correlation in MT.
% \todo{SE: We had sentence-level metrics, but here you mean something related to MT? Wei: updated}
Results reveal that the lexical metric \metric{SentBLEU} can correctly assign lower scores to system translations of low quality, while it struggles in judging system translations of high quality by assigning them lower scores. Our finding agrees with the observations found in \citet{chaganty:2018,novikova-etal-2017-need}: lexical metrics correlate better with human judgments on texts of low quality than high quality. 
\citet{peyrard-2019-studying} further show that lexical metrics %are even not worth trusting 
cannot be trusted because they strongly disagree on high-scoring system outputs.
Importantly, we observe that our word mover metric combining BERT can clearly distinguish texts of two polar qualities.

% Score distribution for one lexical matching metric \metric{SentBLEU} and one semantic metric \metric{Wmd-1-BERT} On WMT17 German-to-English dataset. \metric{SentBLEU} correctly assign low scores to low quality translations, while it struggles in judging high quality ones. Whereas, \metric{Wmd-1-BERT} can clearly distinguish both cases.

%  that while ROUGE is able to correctly assign low scores to texts with low quality, it struggles in judging texts with high quality and often assigns them low-scoring ROUGE. This finding agrees with the observation reported in []
% that automatic metrics correlate better with human
% judgments on bad examples than average or good
% examples. However, we find that semantic metrics are...

\begin{figure}
\begin{minipage}{0.49\linewidth}  
	\centerline{\includegraphics[width=\linewidth]{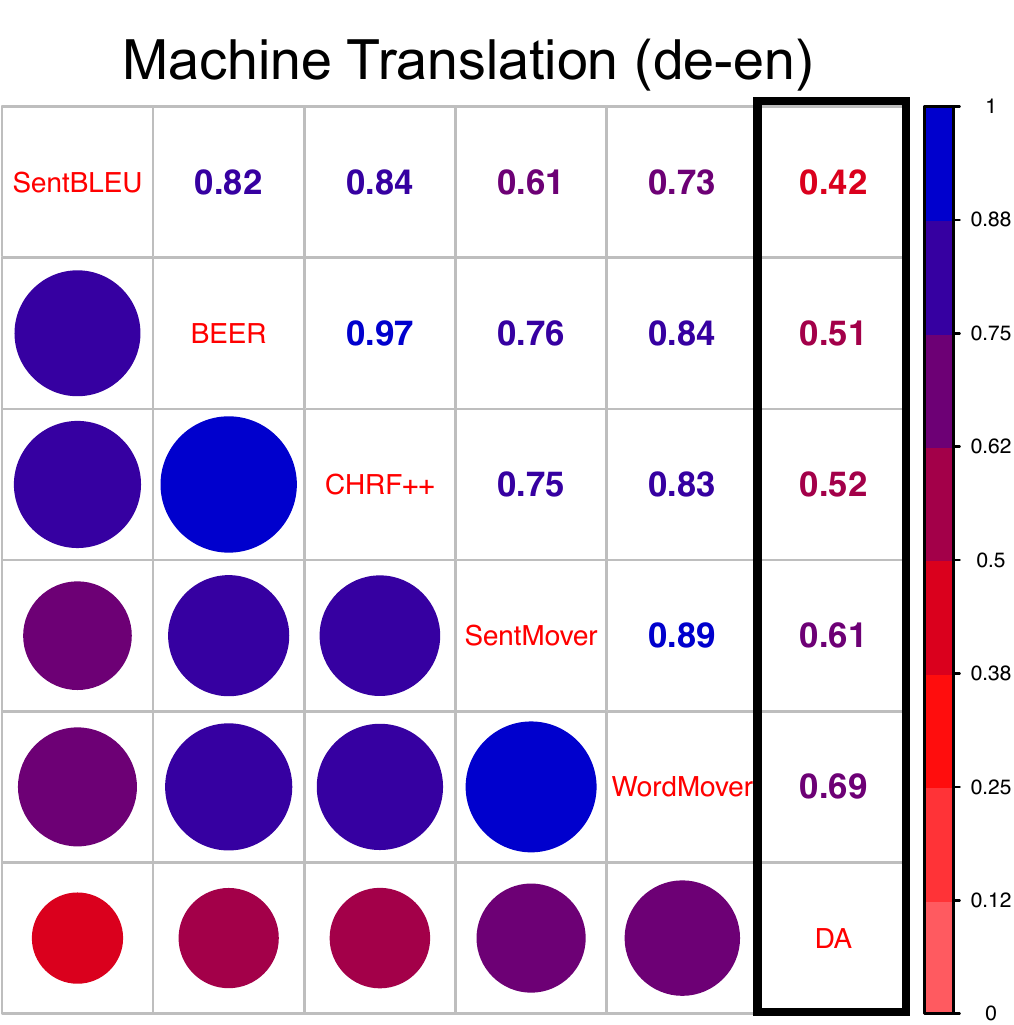}}  
% 	\centerline{(a) this is a placeholder ($n$-gram matching metrics)}  
\end{minipage} 
\begin{minipage}{0.48\linewidth}  
	\centerline{\includegraphics[width=\linewidth]{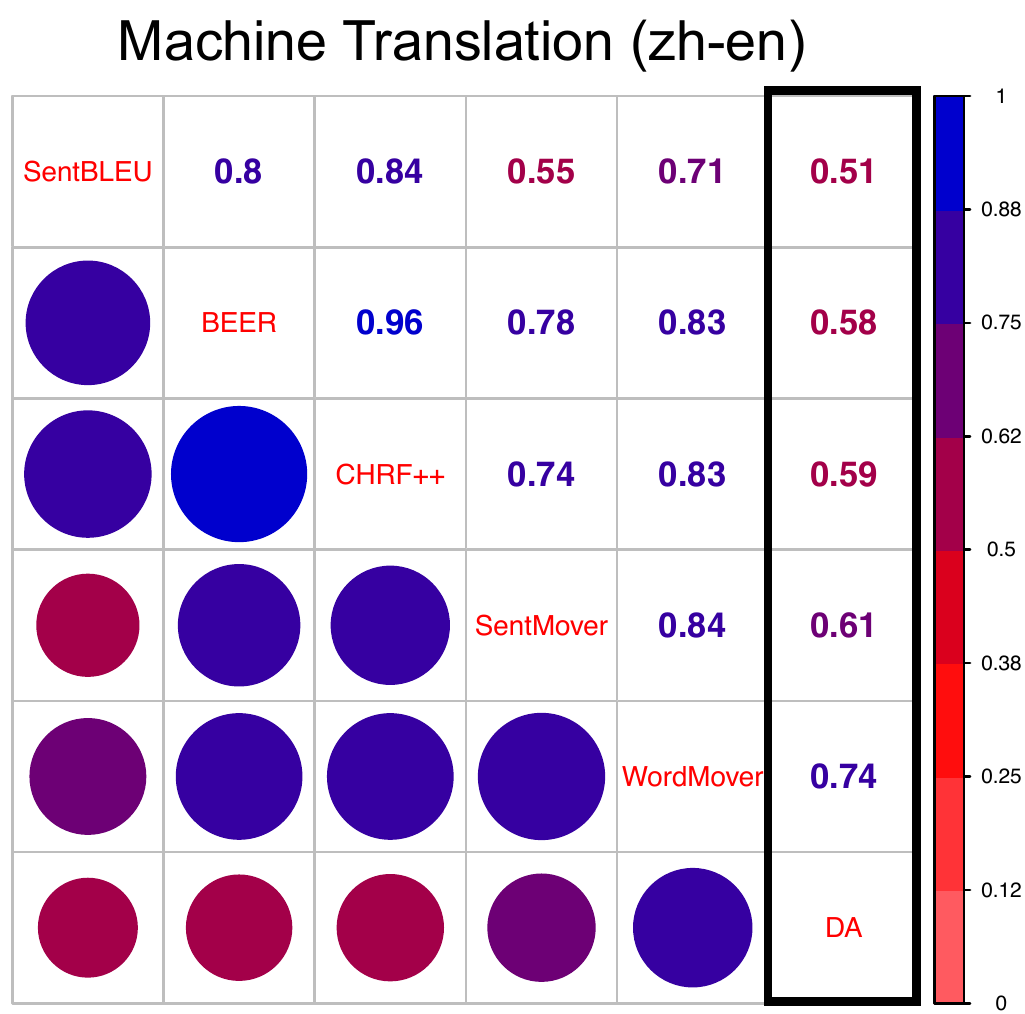}}  
% 	\centerline{(b) this is a placeholder (semantic metrics)}  
\end{minipage} 
 \caption{
 Correlation in similar language (de-en) and distant language (zh-en) pair, where bordered area shows correlations between human assessment and metrics, the rest shows inter-correlations across metrics and DA is direct assessment rated by language experts.
%  [SE: say that this is twice the same information. What is DA? Btw. I thought your metric is called MoverScore, but the graphic says WordMover. Which WordMover variant is it?]
 }
 \label{fig:color_example}  
\vspace{-0.15in}
\end{figure} 

\begin{figure}[t]
\centering
\includegraphics[width=.6\linewidth]{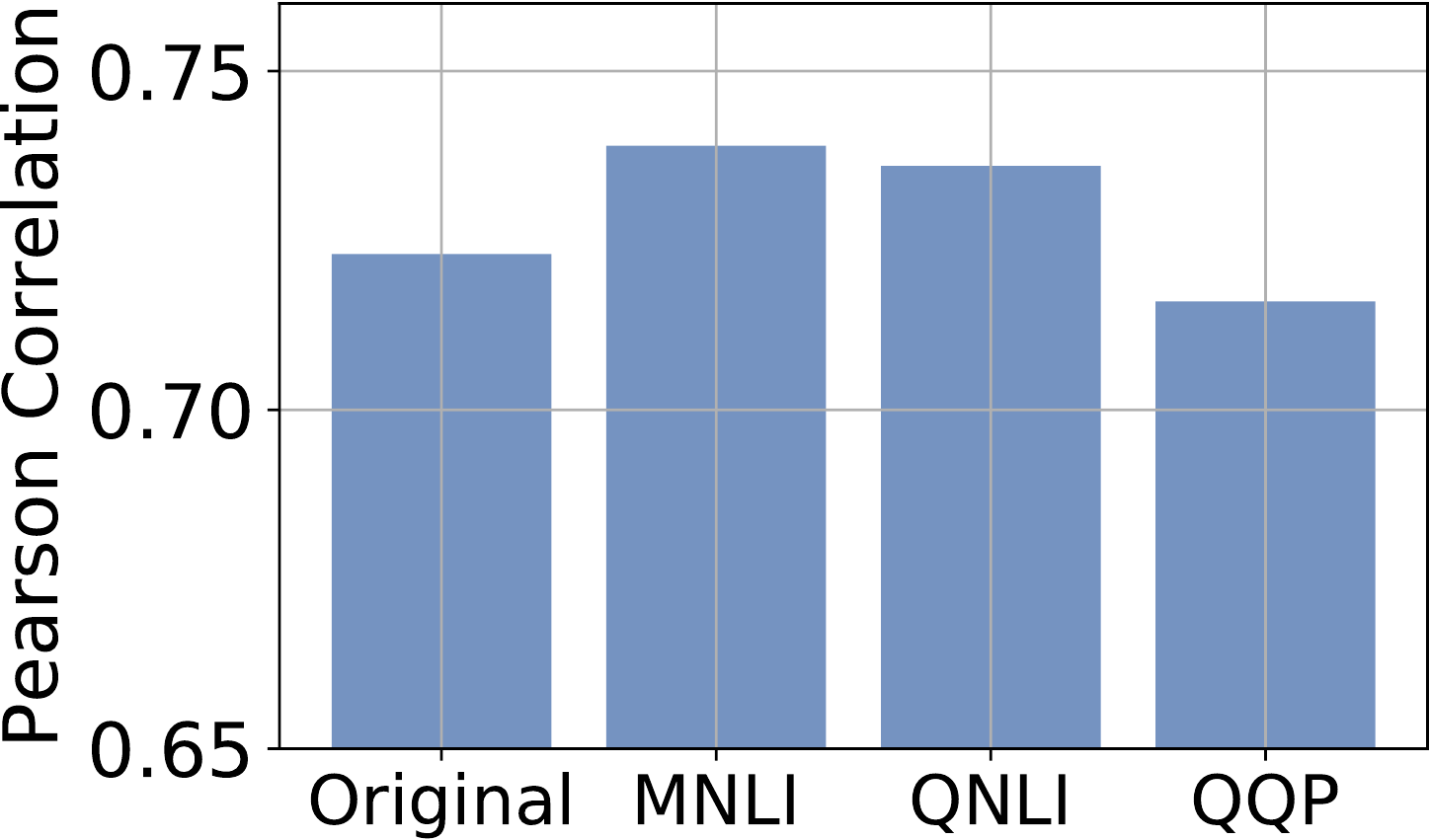} %
\caption{Correlation is averaged over 7 language pairs.}
 \label{fig:task-transfer-analysis}  
\vspace{-0.05in}
\end{figure}

\vspace{0.1in}
\noindent\textbf{Correlation Analysis}\quad
In Figure \ref{fig:color_example}, we observe existing metrics for MT evaluation attaining medium  correlations (0.4-0.5) with human judgments but high inter-correlations between themselves.
% but they perform poorly at distinguishing translations that fall on two extremes.
% of two polar qualities. 
In contrast, 
%we observe that 
% our Mover Distance metric variants, 
our metrics %, the Word and Sentence Mover combining BERT, 
can attain high correlations (0.6-0.7) with human judgments, performing robust across different language pairs. 
%\todo{SE: with what? humans? Wei:updated} 
We believe that our improvements come from clearly distinguishing translations that fall on two extremes.
\vspace{0.1in}
\noindent\textbf{Impact of Fine-tuning Tasks}\quad
Figure \ref{fig:task-transfer-analysis} compares Pearson correlations with our word mover metrics combining BERT fine-tuned on three different tasks. We observe that fine-tuning on closely related tasks improves correlations, especially fine-tuning on MNLI leads to an impressive improvement by 1.8 points on average.

\subsection{Discussions}
% We study our evaluation metrics with a suite of combinations:(i) the granularity of embedding, 
% Firstly, we study the granularity of embeddings and 
We showed that our metric combining contextualized %word 
embeddings and 
%word 
Earth Mover's Distance outperforms strong unsupervised metrics on 3 out of 4 tasks, i.e., METEOR++ on machine translation by 5.7 points, SPICE on image captioning by 3.0 points, and METEOR on dialogue response generation by 2.2 points. The best correlation we achieved is combining contextualized word embeddings and WMD, which even rivals or exceeds SOTA task-dependent \emph{supervised} metrics across  different tasks. Especially in machine translation, our word mover metric pushes correlations in machine translation to 74.3 on average (5.8 points over the SOTA supervised metric and 2.4 points over contemporaneous BERTScore). The major improvements come from contextualized BERT embeddings rather than word2vec and ELMo, and from fine-tuning BERT on large NLI datasets. However, we also observed that soft alignments (MoverScore) marginally outperforms hard alignments (BERTScore). 
% ---%even though soft alignments take more effort in finding many-to-one mappings. 
% those are more costly to obtain, however. \todo{Wei: consider to delete this time-costly point or compare them using time complexity}
Regarding the effect of $n$-grams in word mover metrics, 
%we observed %that our metric with unigram-embeddings is slightly better or on par with bigram-embeddings. 
%that 
unigrams slightly outperforms 
%\todo{outperforms; present sense is used to describe the observations throughout. Wei: updated} 
bigrams on average. 
For the effect of aggregation functions, we suggested 
effective techniques for layer-wise consolidations, namely $p$-means and routing, both of which are close to the performance of the best layer and on par with each other (see the appendix). 
% We will continue to study these problems in future work.

% the effect of n-grams 

% We conduct extensive experiments for 
% In this paper, we considered a very general framework for inducing an unsupervised metric for assessing text generation performance. Our metric uses Word Mover Distance and (contextualized) word embeddings and improves over the most common standard metrics in four different NLG fields: [a,b,c,d], oftentimes very substantially [make an example how we beat up BLEU]. In addition, we explored the effect of aggregation information (min/max/avg improve over avg by xx pp, see appendix) in contextualizers, the effect of n-grams in WMD (we mostly found that $n=1$ is better or on par with $n=2$), the effect of soft distribution matching vs.\ hard alignment (our metric tends to outperform the 
% contemporaneous
% BERTScore). The major improvements come from contextualized word representations rather than static ones, as previously used, and from fine-tuning these on large NLI datasets. 
% We hope that this inspires you guys, and wish you a good night! Cheers from team WEI. 

\label{sec:results}

\section{Conclusion}\label{sec:conclusion}
%In this paper 
We investigated new unsupervised evaluation metrics for text generation systems combining contextualized embeddings with 
Earth Mover's Distance.
% Word Mover's distance. 
%to understand to what extent the system-generated text has deviated from reference texts. 
We experimented with two variants of our metric, sentence mover and word mover. %and 
The latter has demonstrated %promising results and 
strong generalization ability across four text generation tasks, oftentimes even outperforming supervised metrics. 
%of similar nature. 
Our metric provides a promising direction towards a holistic metric for text generation and a direction towards more `human-like' \citep{eger-etal-2019-text} evaluation of text generation systems. 

In future work, we plan to avoid the need for costly human references in the evaluation of text generation systems, and instead base evaluation scores on source texts and system predictions only, which would allow for `next-level', unsupervised (in a double sense) and unlimited evaluation \citep{LouisN13,DBLP:conf/emnlp/no_reference19}. 

%\todo[inline]{SE: The name MoverScore appears nowhere but in the title. Also: sometimes you write "metric" and sometimes "metrics", sometimes n-gram and sometimes $n$-gram. The name "Earth Mover Distance", from the title, appears exactly once. In all other cases, it's called WordMover or Mover. Wei: expose to MoverScore more times}

% \todo{SE: The good impression is mitigated by some ablation test being only performed on certain tasks}
% \todo[inline]{SE:
% In this paper, we considered a very general framework for inducing an unsupervised metric for assessing text generation performance. Our metric uses Word Mover Distance and (contextualized) word embeddings and improves over the most common standard metrics in four different NLG fields: [a,b,c,d], oftentimes very substantially [make an example how we beat up BLEU]. In addition, we explored the effect of aggregation information (min/max/avg improve over avg by xx pp, see appendix) in contextualizers, the effect of n-grams in WMD (we mostly found that $n=1$ is better or on par with $n=2$), the effect of soft distribution matching vs.\ hard alignment (our metric tends to outperform the 
% contemporaneous
% BERTScore). The major improvements come from contextualized word representations rather than static ones, as previously used, and from fine-tuning these on large NLI datasets. 
% We hope that this inspires you guys, and wish you a good night! Cheers from team WEI. 
% }

\section*{Acknowledgments}
We thank the anonymous reviewers for their comments, which greatly improved the final version of the paper. 
This work has been supported by the German Research Foundation as part of the Research Training
Group Adaptive Preparation of Information from Heterogeneous Sources (AIPHES) at the Technische
Universit\"at Darmstadt under grant No. GRK 1994/1.
Fei Liu is supported in part by NSF grant IIS-1909603.
% \clearpage
\bibliography{emnlp-ijcnlp-2019,fei_summ_eval,wei}
\bibliographystyle{acl_natbib}

\appendix
\onecolumn
\section{Supplemental Material}
\label{sec:supplemental}
\subsection{Proof of Prop.~\ref{prop:main}}
% \paragraph{Analysis on BERTScore and WMD}

In this section, we prove Prop.~\ref{prop:main} in the paper about viewing BERTScore (precision/recall) as a (non-optimized) Mover Distance.

% \begin{prop}
% % The precision and recall of BERTScore are the hard forms of Word Mover Distance, and BERTScore F1 is the harmonic-mean of the above two forms.
% % \todo[inline]{SE: I think it's safer to say that BERTScore (precision/recall) admits a (non-optimized) representation $\sum C\cdot F$ where $C$ is a ``distance'' matrix and $F$ is a non-optimized transportation matrix.}
% BERTScore (precision/recall) is a non-optimized Mover Distance $\sum C\cdot F$, where $C$ is a ``distance'' matrix and $F$ is a non-optimized transportation matrix. 
% \end{prop}
As a reminder, the WMD formulation is:
\begin{align*}
& \WMD(\bx^n,\by^n) := \min_{\mF\in\R^{|\bx^n|\times |\by^n|}}\sum_{i,j} \mC_{ij} \cdot \mF_{ij}\\
&\text{s.t. } \1^{\intercal}\mF^{\intercal}\1=1,\;\;\1^{\intercal}\mF\1=1.
\end{align*}
where $\mF^{\intercal}\1=\f_x^n$ and $\mF\1=\f_y^n$. Here, $\f_x^n$ and $\f_y^n$ denote vectors of weights for each $n$-gram of $\bx^n$ and $\by^n$.

%Consider $n=1$, then BERTScore can be formulated as:
BERTScore is defined as: 
\begin{align}
\methodr  &= \frac{\sum_{y_i^1 \in \by^1} \idf(y_i^1)  \max_{x_j^1 \in \bx^1} E(x_j^1)^\intercal E(y_i^1)}{\sum_{y_i^1\in \by^1} \idf(y_i^1)} \nonumber \\
\methodp  &= \frac{\sum_{x_j^1 \in \bx^1} \idf(x_j^1)  \max_{y_i^1 \in \by^1} E(y_i^1)^\intercal E(x_j^1)}{\sum_{x_j^1\in \bx^1} \idf(x_j^1)} \nonumber \\
\methodf &= 2\frac{\methodp \cdot \methodr }{ \methodp + \methodr }\;\;. \nonumber
\end{align}
%If 
%Let $C$ be defined as: %are considered as:
%\begin{align*}
%%&F_{ij} \sim U(0,1) ,\;\; \text{s.t. }
%%\1^{\intercal}F^{\intercal}\1=1,\;\;\1^{\intercal}F\1=1\\
%&C_{ij}= \frac{\idf(y_i^1)\max_{x_j^1 \in \bx^1} %E(x_j^1)^\intercal E(y_i^1)}{\sum_{y_i^1\in \by^1} \idf(y_i^1)}
%\end{align*}
%where $F$ is a uniform transportation cost matrix, and $C$ is a %distance matrix. 
Then, $\methodr$ can be formulated in a ``quasi'' WMD form:

\begin{align*}
& \methodr(\bx^1,\by^1) := \sum_{i,j} \mC_{ij} \cdot \mF_{ij}\\
&\mF_{ij}=\begin{cases}
      \frac{1}{M} & 
      %i,j=\argmax_{y_i^1\in \by^1}E(x_i^1)^\intercal E(y_i^1)\\ 
      \text{if } x_j=\argmax_{\hat{x}_j^1\in \bx^1}E(y_i^1)^\intercal E(\hat{x}_j^1)\\ 
      0 & \text{otherwise}
\end{cases}\\
&\mC_{ij}=\begin{cases}
      \frac{M}{Z}\idf(y_i^1) E(x_j^1)^\intercal E(y_i^1) & \text{if } x_j=\argmax_{\hat{x}_j^1\in \bx^1}E(y_i^1)^\intercal E(\hat{x}_j^1)\\ 
      0 & \text{otherwise}
\end{cases}\\
\end{align*}
where $Z=\sum_{y_i^1 \in \by^1}\idf(y_i^1)$ and $M$ is the size of $n$-grams in $\bx^1$. Similarly, we can have $\methodp$ in a quasi WMD form (omitted). Then, $\methodf$ can be formulated as harmonic-mean of two WMD forms of $\methodp$ and $\methodr$.

% Intuitively, the standard \WMD allows several semantically related words in one system output travel to each word in the respective reference, whereas in BERTScore, each word in a reference only receives the incoming travel cost from the most similar word in the system output and in reverse. Therefore, BERTScore can be formulated as unigram \WMD with the specific $\mC$ and $\mF$ (need to be revised).

\subsection{Routing}
In this section, we study the aggregation function $\phi$ with a routing scheme, which has achieved good results in other NLP tasks \cite{zhao:2018, zhao-etal-2019-towards}.
Specifically, we introduce a nonparametric clustering with Kernel Density Estimation (KDE) for routing since KDE bridges a family of kernel functions with underlying empirical distributions, which often leads to computational efficiency \cite{zhangfast}, defined as:
% In this paper, we employ routing to output aggregated representations from contextualizers, however, we notice that existing routing algorithms do not consider the adaptive iterations for different examples.
% during clustering since they employ a fixed number of iterations for all examples. To address the issue, 
% Therefore, we introduce an adaptive routing algorithm upon kernel density estimation [], defined as:
\begin{align*}
&\min_{\bv,\gamma}
% _{V\in\R^{T\times d},\Gamma \in \R^{L\times T}}
% \: \langle V, \Gamma \rangle 
f(\bz)
=\sum_{i=1}^L \sum_{j=1}^T\gamma_{ij}k(d(\bv_j - \bz_{i,j}))\\
% &\min_{\mathbf{V},\Gamma} \: f(\mathbf{V}, \Gamma) =\sum_{i,j}\gamma_{ij}k(\mathbf{v}_j, \mathbf{z}_i)\\
% & \text{where} \;\; k(\mathbf{v}_j, \mathbf{z}_i) = 1- \langle \mathbf{v}_j, \mathbf{z}_i \rangle.
& \textit{s.t.}\quad\forall{i,j}:\gamma_{ij}>0,\sum\limits_{j=1}^{L}\gamma_{ij}=1.
\end{align*}
where $d(\cdot)$ is a distance function, $\gamma_{ij}$ denotes the  underlying closeness between the aggregated vector $\bv_j$ and vector $\bz_i$ in the $i$-th layer, and $k$ is a kernel function. Some instantiations of $k(\cdot)$ \cite{wand:1994} are:
\begin{align*}
\mathit{Gaussian}: k(x)\triangleq\exp{(-\frac{x}{2})},\;\;
\mathit{Epanechnikov}: k(x)\triangleq\left\{\begin{aligned}1-x && x\in{[0,1)}\\ 0 && x\ge{1}.\end{aligned}\right.
\end{align*}
% $$k(x)=
% \begin{cases}
% 1 - x & x \in [0,1)\\
% 0 & x \geq 1
% \end{cases}$$
One typical solution for KDE clustering to minimize $f(\bz)$ is taking Mean Shift \cite{comaniciu:2002}, defined as:
%  typical clustering method based on the framework of KDE
% We adopt the mean shift algorithm~\cite{comaniciu2002mean} to maximize $f(v, c)$:
% Intuitively, a shorter distance means that a sequence of representations $\mathbf{v}_1,...,\mathbf{v}_T\in\mathbb{R}^d$ are aggregated well by our routing. 
\begin{align*}
	\nabla f(\bz)=\sum_{i,j} \gamma_{ij} k'(d(\bv_j, \bz_{i,j}))\frac{\partial{d(\bv_j,  \bz_{i,j})}}{\partial{\bv}} \nonumber
\end{align*}
Firstly, $\bv_j^{\tau+1}$ can be updated while $\gamma_{ij}^{\tau+1}$ is fixed:
\begin{align*}
	\bv_j^{\tau+1} = \frac{\sum_{i} \gamma_{ij}^\tau k'(d(\bv_j^\tau, \bz_{i,j})) \bz_{i,j}}{\sum_{i,j} k'(d(\bv_j^\tau, \bz_{i,j}))}
\end{align*}
Intuitively, $\bv_j$ can be explained as a final aggregated vector from $L$ contextualized layers.
% normalized weighted sum over L layers vectors $\mathbf{z}_i$. 
% The weight consists of the
% coefficient rij , the prior knowledge of candidate sample and the derivate of kernel
% function.
Then, we adopt SGD to update $\gamma_{ij}^{\tau+1}$:
\begin{align*}
	\gamma_{ij}^{\tau+1} = \gamma_{ij}^\tau + \alpha \cdot k(d(\bv_j^\tau, \bz_{i,j}))
\end{align*}
where $\alpha$ is a hyperparameter to control step size. The routing process is summarized in Algorithm \ref{alg:1}.

% $c_{ij}$ is the coupling coefficient that measures the intensity of connection between one ``prediction capsule'' $\hat u_{j|i}$ and aggregated high-level capsule $v_j$, 
\begin{algorithm}
\caption{Aggregation by Routing}\label{alg:1}
\begin{algorithmic}[1]
\small
\State \textbf{procedure} ROUTING($\bz_{ij}$, $\ell$)
\State Initialize $\forall i,j: \gamma_{ij}=0$
\While {true}
  \ForAll{representation $i$ and $j$ in layer $\ell$ and  $\ell+1$}
    % \State 
    $\gamma_{ij} \gets \softmax(\gamma_{ij})$
  \EndFor   
  
  \ForAll{representation $j$ in layer $\ell+1$}
    \State $\bv_j \gets \sum_{i} \gamma_{ij} k'(\bv_j, \bz_i) \bz_i / \sum_{i} k'(\bv_i, \bz_i)$
  \EndFor    

  \ForAll{representation $i$ and $j$ in layer $\ell$ and  $\ell+1$} $\gamma_{ij}\gets \gamma_{ij} + \alpha \cdot k(\bv_j, \bz_i)$
    % \State $\gamma_{ij}\gets \gamma_{ij} + \alpha k(v_j, z_i)$
  \EndFor  

  \State $\rm{loss} \gets \log(\sum_{i,j}\gamma_{ij}k(\bv_j, \bz_i))$
  \If{$|\rm{loss}- \rm{preloss}| < \epsilon$}
      \State \textbf{break}
  \Else
      \State $\rm{preloss} \gets \rm{loss}$
  \EndIf
\EndWhile
\State \textbf{return} $v_j$
\end{algorithmic}
\end{algorithm}

\paragraph{Best Layer and Layer-wise Consolidation}
Table \ref{tab:wmt17-aggregation} compares our word mover based metric combining BERT representations on different layers with stronger BERT representations consolidated from these layers (using $p$-means and routing). We often see that which layer has best performance is task-dependent, and our word mover based metrics (WMD) with $p$-means or routing schema come close to the oracle performance obtained from the best layers.

% We often see the initial layers perform poorly, while using the aggregations of output layers could rival with the performance in best layer.

% Which layer has the best performance is highly task-dependent. One could decide this question for each NLG evaluation task by comparing each layer representation on its ability to correlate with human judgments for the specific task, but
% Meanwhile, which layer leads to the best performance is a mystery and highly task-dependent. While one can simply find the best layer in contextualzers with human provided judgments, it is impractical because one cannot generalize it to a new dataset without human scores. Instead of a lottery-choice from the intermediate layers, we consider aggregating contexualized word vectors coming from different layers into one stronger and richer vector, expecting to reach to the oracle performance obtained from best layers.

\setlength{\tabcolsep}{8.8pt}
\begin{table*}[h!]
    \small    
    \centering
    \begin{tabular}{l ccccccc}
    \toprule
    & \multicolumn{7}{c}{\textbf{Direct Assessment}}\\
    Metrics & cs-en & de-en & fi-en & lv-en & ru-en & tr-en & zh-en \\
    % \midrule
    % \multicolumn{9}{l}{\textit{Semantic distance combining contextual word embedding (Word-WMD)}}\\
    \midrule
    % \metric{Wmd-1 + BERT + Layer 1} & - & - & - & - & - & - & - \\ 
    % \metric{Wmd-1 + BERT + Layer 2} & - & - & - & - & - & - & - \\ 
    % \metric{Wmd-1 + BERT + Layer 3} & - & - & - & - & - & - & - \\ 
    % \metric{Wmd-1 + BERT + Layer 4} & - & - & - & - & - & - & - \\ 
    % \metric{Wmd-1 + BERT + Layer 5} & - & - & - & - & - & - & - \\ 
    % \metric{Wmd-1 + BERT + Layer 6} & - & - & - & - & - & - & - \\ 
    % \metric{Wmd-1 + BERT + Layer 7} & - & - & - & - & - & - & - \\ 
    \metric{Wmd-1 + BERT + Layer 8} & .6361 & .6755 & .8134 & .7033 & .7273 & .7233 & .7175 \\ 
    \metric{Wmd-1 + BERT + Layer 9} & .6510 & .6865 & .8240 & .7107 & .7291 & \textbf{.7357} & \textbf{.7195} \\ 
    \metric{Wmd-1 + BERT + Layer 10} & .6605 & \textbf{.6948} & \textbf{.8231} & .7158 & \textbf{.7363} & .7317 & .7168 \\
    \metric{Wmd-1 + BERT + Layer 11} & \textbf{.6695} & .6845 & .8192 & .7048 & .7315 & .7276 & .7058 \\
    \metric{Wmd-1 + BERT + Layer 12} & .6677 & .6825 & .8194 & \textbf{.7188} & .7326 & .7291 & .7064 \\
    \midrule
    % \multicolumn{9}{l}{\textit{Semantic metrics via aggregated word embedding}}\\
    % \midrule
    \metric{Wmd-1 + BERT + Routing} & .6618 & .6897 & .8225 & .7122 & .7334 & .7301 & .7182 \\ 
    \metric{Wmd-1 + BERT + PMeans} & .6623 & .6873 & .8234 & .7139 & .7350 & .7339 & .7192 \\ 
    % \metric{Wmd-1 + BERT + Average over Last 5 layers} & .6504 & .6823 & - & - & - & - & - \\ 
    % \metric{Wmd-1 + BERT + Layer 8} & 0.651 & 0.686 & 0.823 & 0.710 & 0.729 & 0.729 & 0.720 & 0.721 \\ 
    % % \metric{BERT-2-Wmd + best layer} & 0.665 & 0.688 & 0.821 & 0.712 & 0.728 & 0.735 & 0.719 & 0.724 \\ 
    % \midrule
    % \metric{Wmd-1 + BERTPmeans} & - & - & - & - & - & - & - & - \\ 
    % % \metric{Wmd-2 + BERTPmeans} & - & - & - & - & - & - & - & - \\ 
    % \metric{Wmd-1 + BERTRouting} & 0.658 & 0.689 & \textbf{0.823} & 0.712 & 0.733 & 0.730 & 0.718 & 0.723 \\ 
    % \metric{Wmd-2 + BERTRouting} & \textbf{0.673} & \textbf{0.691} & 0.819 & \textbf{0.713} & \textbf{0.735} & 0.732 & 0.715 & \textbf{0.725} \\ 
    % \midrule
	\hline  
    \end{tabular}
    % }
    \caption{Absolute Pearson correlations with segment-level human judgments on WMT17 to-English translations.}
    \label{tab:wmt17-aggregation}
\end{table*}

\paragraph{Experiments}
Table ~\ref{tab:wmt17-all}, \ref{tab:tac-all} and \ref{tab:dialogue-all} show correlations between metrics (all baseline metrics and word mover based metrics) and human judgments on machine translation, text summarization and dialogue response generation, respectively. We find that word mover based metrics combining BERT fine-tuned on MNLI %performing  best,
have highest correlations with humans, 
%which 
%surpassing 
outperforming
all of the unsupervised metrics and even %outperform 
supervised metrics like $\metric{RUSE}$ and $\metric{$S^{3}_{full}$}$.  %but 
%The performance gains brought by 
Routing and $p$-means %are generally comparable.
perform roughly equally well. 
% on machine translation and text summarization.

\setlength{\tabcolsep}{5pt}
\begin{table}[h!]
    \footnotesize    
    \centering
    \begin{tabular}{c|l|ccccccc c}
    \toprule
    & & \multicolumn{8}{c}{\textbf{Direct Assessment}}\\
    Setting & Metrics & cs-en & de-en & fi-en & lv-en & ru-en & tr-en & zh-en & Average \\
    % \midrule
    % \multicolumn{10}{l}{\textit{Learning Metrics trained from human judgments}}\\
    \midrule
    \multirow{5}{*}{\metric{Baselines}}
    & \metric{BLEND} & 0.594 & 0.571 & 0.733 & 0.594 & 0.622 & 0.671 & 0.661 & 0.635 \\
    & \metric{RUSE} & 0.624 & 0.644 & 0.750 & 0.697 & 0.673 & 0.716 & 0.691 & 0.685 \\
    % \midrule
    % \multicolumn{10}{l}{\textit{N-gram metrics upon word-overlapping}}\\
    % \midrule
    % \multirow{4}{*}{Unsupervised} 
    % & \metric{BLEU-4} & 0.330 & 0.367 & 0.492 & 0.321 & 0.348 & 0.462 & 0.459 & 0.397\\
    & \metric{SentBLEU} & 0.435 & 0.432 & 0.571 & 0.393 & 0.484 & 0.538 & 0.512 & 0.481 \\
    & \metric{chrF++} & 0.523 & 0.534 & 0.678 & 0.520 & 0.588 & 0.614 & 0.593 & 0.579 \\
    & \metric{METEOR++} & 0.552 & 0.538 & 0.720 & 0.563 & 0.627 & 0.626 & 0.646 & 0.610 \\
    & \metric{BERTScore-F1} & 0.670 & 0.686 & 0.820 & 0.710 & 0.729 & 0.714 & 0.704 & 0.719 \\
    \midrule
    % \multicolumn{10}{l}{\textit{Semantic metrics upon word embedding}}\\
    % \midrule
    \multirow{7}{*}{\metric{Word-Mover}}
    
    & \metric{Wmd-1 + W2V} &0.392 & 0.463 & 0.558 & 0.463 & 0.456 & 0.485 & 0.481 & 0.471 \\
    % & \metric{BERTScore-F1} & \textbf{0.618} & \textbf{0.665} & 0.803 & 0.684 & 0.697 & 0.701 & 0.701 & - \\
    % & \metric{Wmd-1 + ELMO + Routing} & 0.540 & 0.556 & 0.734 & 0.524 & 0.575 & 0.644 & 0.630 & 0.600 \\
    %  & \metric{BERTScore-F1} & 0.655 & 0.681 & 0.821 & 0.712 & 0.725 & 0.717 & 0.712 & 0.717 \\
    % the numbers above are slightly different from the paper numbers below
    % & 0.670 & 0.686 & 0.820 & 0.710 & 0.729 & 0.714 & 0.704 & 0.719 \\
    % & \metric{WMD-1 + BERT} & 0.651 & 0.686 & 0.823 & 0.710 & 0.729 & 0.729 & \textbf{0.720} & 0.721 \\ 
    % & \metric{WMD-2 + BERT} & 0.665 & 0.688 & 0.821 & 0.712 & 0.728 & 0.735 & 0.719 & 0.724 \\ 
    & \metric{Wmd-1 + BERT + Routing} & 0.658 & 0.689 & 0.823 & 0.712 & 0.733 & 0.730 & 0.718 & 0.723 \\ 
    % & \metric{Wmd-2 + BERT + Routing} & 0.673 & 0.691 & 0.819 & 0.713 & 0.735 & 0.732 & 0.715 & 0.725 \\ 
    % \midrule
    % \multirow{2}{*}{Pretraining-MNLI} 
    %---------------------BERT+Infersent-------------------------
    % & \metric{BERTScore-F1} & 0.624 & 0.672 & 0.805 & \textbf{0.705} & 0.695 & 0.713 & 0.701 & - \\
    % & \metric{BERTRouting-F1} & \textbf{0.631} & \textbf{0.682} & \textbf{0.810} & 0.702 & \textbf{0.695} & \textbf{0.732} & \textbf{0.706} & - \\
    %------------------------------------------------------------
    % & \metric{BERTRouting-F1} & \textbf{0.641} & \textbf{0.690} & \textbf{0.820} & \textbf{0.714} & \textbf{0.707} & \textbf{0.732} & \textbf{0.717} & -  \\
    & \metric{Wmd-1 + BERT + MNLI + Routing} & 0.665 & 0.705 & 0.834 & 0.744 & 0.735 & 0.752 & 0.736 & 0.739 \\ 
    & \metric{Wmd-2 + BERT + MNLI + Routing} & 0.676 & 0.706 & 0.831 & 0.743 & 0.734 & 0.755 & 0.732 & 0.740 \\ 
    
    &\metric{Wmd-1 + BERT + PMeans} & 0.662 & 0.687 & 0.823 & 0.714 & 0.735 & 0.734 & 0.719 & 0.725\\ 
    % &\metric{Wmd-1 + BERT + Best Layer} & 0.669 & 0.695 & 0.824 & 0.719 & 0.736 & 0.735 & 0.720 & 0.728\\ 
    
    % & 0.686 & 0.823 & 0.710 & 0.729 & 0.729 & 0.720 & 0.721 \\ 
    % &\metric{Wmd-2 + BERT + Best Layer} & 0.679 & 0.697 & 0.820 & 0.721 & 0.739 & 0.737 & 0.724 & 0.731 \\ 
    % & 0.688 & 0.821 & 0.712 & 0.728 & 0.735 & 0.719 & 0.724 \\
    &\metric{Wmd-1 + BERT + MNLI + PMeans} & 0.670 & 0.708 & \textbf{0.835} & \textbf{0.746} & \textbf{0.738} & 0.762 & \textbf{0.744} & \textbf{0.743}\\ 
    
    &\metric{Wmd-2 + BERT + MNLI + PMeans} & \textbf{0.679} & \textbf{0.710} & 0.832 & 0.745 & 0.736 & \textbf{0.763} & 0.740 & \textbf{0.743}\\ 

	\hline  
    \end{tabular}
    % }
    \caption{Absolute Pearson correlations with segment-level human judgments on WMT17 to-English translations.}
    \label{tab:wmt17-all}
\end{table}

% 48 and 44 topics, respectively, and each topic consists of 10 news articles, 1 system summary from each summarization system and 4 human reference summaries. Both datasets provide two summary-level human scores (Pyramid and overall responsiveness), which assign 
% \begin{table}[h!]
% \scriptsize
% 	\centering
%     	\resizebox{1\columnwidth}{!}{
	
% 	\begin{tabular}{|l|c c c | c c c |}
% 		\toprule
%         &\multicolumn{3}{c|}{\textbf{Pyramid}}
%         &\multicolumn{3}{c|}{\textbf{Responsiveness}}
%          \\
% \cmidrule{2-4}\cmidrule{5-7}
% \textbf{Metric} & \textbf{Kendall} & \textbf{Pearson}& \textbf{Spearman} & \textbf{Kendall} & \textbf{Pearson} & \textbf{Spearman} \\
% \midrule
% \multicolumn{7}{|l|}{Word-Overlap Metrics}\\
% \midrule
% ROUGE-1 & 48.4 & 76.6 & 65.4 & - & - & -\\
% ROUGE-2 & 48.0 & 76.3 & 65.3 & - & - & -\\
% ROUGE-L & 49.5 & 75.6 & 66.6 & - & - & -\\
% \midrule
% \multicolumn{7}{|l|}{Semantic Metrics (Supervised Sequence Model)}\\
% \midrule
% InferSent & 48.2 & 75.9 & 65.3 & - & - & - \\
% BERT-base & \textbf{51.3} & \textbf{77.6} & \textbf{69.1} & - & - & - \\
% 	\bottomrule	
% 	\end{tabular}
% 	}
%     \caption{Experimental results on TAC-2008}
%     \label{tab:5}
%     \vspace{-0.5cm}
% \end{table}
\setlength{\tabcolsep}{4.5pt}
\begin{table}
		\small
		\centering
  		\begin{tabular}{c|l |cc|cc|cc|cc}
  		\toprule
 		&&\multicolumn{4}{c|}{TAC-2008}	&	\multicolumn{4}{c}{TAC-2009}\\
 		&&\multicolumn{2}{c|}{\textbf{Responsiveness}}&\multicolumn{2}{c|}{\textbf{Pyramid}} &\multicolumn{2}{c|}{\textbf{Responsiveness}} & \multicolumn{2}{c}{\textbf{Pyramid}}\\
 		Setting & Metrics	&	$r$ & $\rho$& $r$   & $\rho$ & $r$ & $\rho$ & $r$   &$\rho$\\
 		\midrule
 		
%  		\multicolumn{10}{l}{\textit{Learning Metrics trained from human judgments}}\\
%         \midrule
        \multirow{13}{*}{\metric{Baselines}}
        &\metric{$S^{3}_{full}$} & 0.696 & 0.558 & 0.753 & 0.652 & 0.731 & 0.552 & 0.838 & 0.724 \\
		&\metric{$S^{3}_{best}$} & 0.715 & 0.595 & 0.754 & 0.652 & 0.738 & \textbf{0.595} & \textbf{0.842} & \textbf{0.731} \\
        
        % \midrule
        % \multicolumn{10}{l}{\textit{N-gram metrics upon word-overlapping}}\\
        % \midrule
        % \multirow{9}{*}{Unsupervised}
		&\metric{TF$*$IDF-1}	& 0.176 & 0.224 & 0.183 & 0.237 & 0.187 & 0.222 & 0.242 & 0.284\\
		&\metric{TF$*$IDF-2}	& 0.047 & 0.154 & 0.049 & 0.182 & 0.047 & 0.167 & 0.097 & 0.233\\
		&\metric{ROUGE-1} & 0.703 & 0.578 & 0.747 & 0.632 & 0.704 & 0.565 & 0.808 & 0.692 \\
		&\metric{ROUGE-2} & 0.695 & 0.572 & 0.718 & 0.635 & 0.727 & 0.583 & 0.803 & 0.694 \\
		&\metric{ROUGE-1-WE}	& 0.571 & 0.450 & 0.579 & 0.458 & 0.586 & 0.437 & 0.653 & 0.516 \\
		&\metric{ROUGE-2-WE}	& 0.566 & 0.397 & 0.556 & 0.388 & 0.607 & 0.413 & 0.671 & 0.481 \\
		&\metric{ROUGE-L} & 0.681 & 0.520 & 0.702 & 0.568 & 0.730 & 0.563 & 0.779 & 0.652 \\
		&\metric{Frame-1} & 0.658 & 0.508 & 0.686 & 0.529 & 0.678 & 0.527 & 0.762 & 0.628 \\
		&\metric{Frame-2} & 0.676 & 0.519 & 0.691 & 0.556 & 0.715 & 0.555 & 0.781 & 0.648 \\
        & \metric{BERTScore-F1} & 0.724 & 0.594 & 0.750 & 0.649 & 0.739 & 0.580 & 0.823 & 0.703 \\
		\midrule
% % 		Pyramid				& .7030 & .6604 & .8528 & --- & --- & --- & 						.7152 & .6386 & .8520 & --- & --- & ---  \\
    % \multicolumn{10}{l}{\textit{Semantic metrics upon word embedding}}\\
    % \midrule
    \multirow{7}{*}{\metric{Word-Mover}}
    
    & \metric{Wmd-1 + W2V} & 0.669 & 0.559 & 0.665 & 0.611 & 0.698 & 0.520 & 0.740 & 0.647 \\
    % & \metric{Wmd-1 + ELMO + Routing} & 0.707 & 0.554 & 0.727 & 0.601 & 0.736 & 0.553 & 0.813 & 0.672 \\
    % & \metric{WMD-1 + BERT} & \textbf{0.731} & 0.601 & 0.755 & 0.662 & 0.745 & \textbf{0.587} & 0.826 & 0.694 \\
    % & \metric{WMD-2 + BERT} & 0.725 & 0.580 & 0.739 & 0.632 & 0.746 & 0.576 & 0.813 & 0.675 \\
    & \metric{Wmd-1 + BERT + Routing} & 0.729 & 0.601 & 0.763 & 0.675 & 0.740 & 0.580 & 0.831 & 0.700 \\
    % & \metric{Wmd-2 + BERT + Routing} & 0.726 & 0.582 & 0.749 & 0.651 & 0.747 & 0.575 & 0.822 & 0.685 \\
    % \midrule
    % \multirow{2}{*}{Pretraining-MNLI} 
    & \metric{Wmd-1 + BERT + MNLI + Routing} & 0.734 & \textbf{0.609} & \textbf{0.768} & \textbf{0.686} & 0.747 & 0.589 & 0.837 & 0.711 \\
    & \metric{Wmd-2 + BERT + MNLI + Routing} & 0.731 & 0.593 & 0.755 & 0.666 & 0.753 & 0.583 & 0.827 & 0.698 \\
    &\metric{Wmd-1 + BERT + PMeans} & 0.729 & 0.595 & 0.755 & 0.660 & 0.742 & 0.581 & 0.825 & 0.690 \\
    &\metric{Wmd-1 + BERT + MNLI + PMeans} & \textbf{0.736} & 0.604 & 0.760 & 0.672 & \textbf{0.754} & 0.594 & 0.831 & 0.701 \\
    &\metric{Wmd-2 + BERT + MNLI + PMeans} & 0.734 & 0.601 & 0.752 & 0.663 & 0.753 & 0.586 & 0.825 & 0.694 \\
		\hline  
		\end{tabular}
 		\caption{Correlation of automatic metrics with summary-level human judgments for TAC-2008 and TAC-2009.}
 		\label{tab:tac-all}
	\end{table}
\vspace{-0.1in}

\setlength{\tabcolsep}{8pt}
\begin{table}
    \small
	\centering
	
	\begin{tabular}{c | l|c c c|c c c}
		\hline\noalign{\smallskip}
		&& \multicolumn{3}{c}{BAGEL} & \multicolumn{3}{c}{SFHOTEL} \\
        Setting & Metrics & \textbf{Inf} & \textbf{Nat} & \textbf{Qual} & \textbf{Inf} & \textbf{Nat} & \textbf{Qual} \\
		\hline\noalign{\smallskip}
        % \multicolumn{7}{l}{\textit{N-gram metrics upon word-overlapping}}\\
        % \midrule
        \multirow{8}{*}{\metric{Baselines}}
        &\metric{BLEU-1} & 0.225 & 0.141 & 0.113 & 0.107 & 0.175 & 0.069 \\
        &\metric{BLEU-2} & 0.211 & 0.152 & 0.115 & 0.097 & 0.174 & 0.071 \\
        &\metric{BLEU-3} & 0.191 & 0.150 & 0.109 & 0.089 & 0.161 & 0.070 \\
        &\metric{BLEU-4} & 0.175 & 0.141 & 0.101 & 0.084 & 0.104 & 0.056 \\
        &\metric{ROUGE-L} & 0.202 & 0.134 & 0.111 & 0.092 & 0.147 & 0.062 \\
        &\metric{NIST} & 0.207 & 0.089 & 0.056 & 0.072 & 0.125 & 0.061 \\
        &\metric{CIDEr} & 0.205 & 0.162 & 0.119 & 0.095 & 0.155 & 0.052 \\
        &\metric{METEOR} & 0.251 & 0.127 & 0.116 & 0.111 & 0.148 & 0.082 \\
        & \metric{BERTScore-F1} & 0.267 & 0.210 & \textbf{0.178} & 0.163 & 0.193 & 0.118 \\
    % \midrule
    % \multicolumn{7}{l}{\textit{Semantic metrics upon word embedding}}\\
    % \midrule
    \midrule
    \multirow{7}{*}{\metric{Word-Mover}}
    & \metric{Wmd-1 + W2V} & 0.222 & 0.079 & 0.123 & 0.074 & 0.095 & 0.021 \\
    % & \metric{WMD-1 + ELMO} & 0.230 & 0.171 & 0.134 & 0.117 & 0.191 & 0.100 \\
    % & \metric{WMD-1 + BERT} & 0.291 & 0.205 & 0.154 & 0.192 & 0.235 & 0.142 \\
    % & \metric{WMD-2 + BERT} & 0.290 & 0.207 & 0.156 & 0.185 & 0.237 & 0.145 \\
    & \metric{Wmd-1 + BERT + Routing} & 0.294 & 0.209 & 0.156 & 0.208 & 0.256 & 0.178 \\
    % & \metric{WMD-2 + BERT + Routing} & 0.293 & 0.210 & 0.155 & 0.200 & 0.238 & 0.150 \\
    % \midrule
    % \multirow{2}{*}{Pretraining-MNLI} 
    & \metric{Wmd-1 + BERT + MNLI + Routing} & 0.278 & 0.180 & 0.144 & \textbf{0.211} & 0.252 & 0.175\\
    & \metric{Wmd-2 + BERT + MNLI + Routing} & 0.279 & 0.182 & 0.147 & 0.204 & 0.252 & 0.172 \\ 
    &\metric{Wmd-1 + BERT + PMeans} & \textbf{0.298} & \textbf{0.212} & 0.163 & 0.203 & 0.261 & 0.182  \\
    &\metric{Wmd-1 + BERT + MNLI + PMeans} & 0.285 & 0.195 & 0.158 & 0.207 & \textbf{0.270} & \textbf{0.183}  \\
    &\metric{Wmd-2 + BERT + MNLI + PMeans} & 0.284 & 0.194 & 0.156 & 0.204 & \textbf{0.270} & 0.182  \\
		\hline  
	\end{tabular}
	\caption{Spearman correlation with utterance-level human judgments for BAGEL and SFHOTEL datasets.}
	\label{tab:dialogue-all}
\end{table}
\setlength{\tabcolsep}{1.4pt}

\end{document}